%% file: joint_pass_good.tex
\documentclass[journal,twocolumn]{IEEEtran}
\usepackage{cite}
\usepackage{amsmath,amssymb,amsfonts}
\usepackage{algorithm}
\usepackage[noend]{algpseudocode}
\usepackage{tikz}
\usetikzlibrary{patterns,matrix,arrows}
\usepackage{graphicx}
\usepackage{textcomp}
\usepackage{xcolor}
\usepackage{url}
\usepackage{tikz}
\usepackage{verbatim}
\usepackage{pgfplots}
\usepackage{subcaption}
\usepackage{graphicx}
\usepackage[inline]{enumitem}
\DeclareMathOperator*{\argmax}{arg\,max} 
 
\pgfplotsset{compat=1.14}

\begin{document}
	\title{FlexPool: A Distributed Model-Free Deep Reinforcement Learning Algorithm for Joint Passengers \& Goods Transportation}
	\author{Kaushik Manchella, Abhishek K. Umrawal, and Vaneet Aggarwal\thanks{The authors are with the School of Industrial Engineering, Purdue University, West Lafayette, IN 47907, email: \{kmanchel,aumrawal,vaneet\}@purdue.edu. V. Aggarwal is also with the School of Electrical and Computer Engineering at Purdue University. }}

	\maketitle
	
\input{abstract}
\input{intro}

\input{statement}

\input{framework}

\input{simulator}
\input{evaluations}

\input{conclusions}
\section{Acknowledgement}
The authors would like to thank Intel for giving us access to the Intel DevCloud cluster for this project. This work was supported in part by  DARPA SAIL-ON project ``Generating Novelty in Open-world Multi-agent Environments (GNOME)". 
	\bibliographystyle{ieeetr}
	\bibliography{references}

	\appendices

\input{ETA}

		\input{notations}
		\section{Table of Notations}
	Table \ref{tbl_not} provides the summary of key notations that are used in this paper. 
	

\end{document}

%% file: abstract.tex
\begin{abstract}

The growth in online goods delivery is causing a dramatic surge in urban vehicle traffic from last-mile deliveries. On the other hand, ride-sharing has been on the rise with the success of ride-sharing platforms and increased research on using autonomous vehicle technologies for routing and matching. The future of urban mobility for passengers and goods relies on leveraging new methods that minimize operational costs and environmental footprints of transportation systems. 

This paper considers combining passenger transportation with goods delivery to improve vehicle-based transportation. %
We propose FlexPool: a distributed model-free deep reinforcement learning algorithm that jointly serves passengers \& goods workloads by learning optimal dispatch policies from its interaction with the environment. %
  The proposed algorithm pools passengers for a ride-sharing service and delivers goods using a multi-hop transit method. These flexibilities decrease the fleet's operational cost and environmental footprint while maintaining service levels for passengers and goods. The dispatching algorithm based on deep reinforcement learning is integrated with an efficient matching algorithm for passengers and goods. Through simulations on a realistic multi-agent urban mobility platform, we demonstrate that FlexPool outperforms other model-free settings in serving the demands from passengers \& goods. FlexPool achieves  30\% higher fleet utilization and 35\% higher fuel efficiency in comparison to (i) model-free approaches where vehicles transport a combination of passengers \& goods without the use of multi-hop transit, and (ii) model-free approaches where vehicles exclusively transport either passengers or goods.

\end{abstract}

\begin{IEEEkeywords}
Ride-sharing, Urban Delivery, Vehicle Dispatch, Deep Q-Network, Reinforcement Learning, Intelligent Transportation, Fleet Management%
\end{IEEEkeywords}

%% file: intro.tex
\section{Introduction}

\subsection{Motivation}

While ride-sharing or ride-splitting has risen to a common service due to its various benefits (for customers, drivers, and sustainability), various forms of crowd-sourced delivery are also on the upsurge in adoption and demand including last mile delivery services like Amazon Flex, urban package delivery services for food like Doordash, and groceries delivery services like Instacart, Shipt, etc. \cite{asdecker2020drives}.    

E-commerce has seen a double digit growth over the past few years due to increasing internet penetration, smartphone adoption, etc. \cite{arora2019m,jo2019impact}. 
As a result of this growth, the deliveries from online orders of goods and services are playing a more significant role in urban transportation systems. The customer seeking convenience has led to increased demands for last-mile delivery \cite{bubner2014logistics}. As a result of the increased demands, from a city transportation perspective, the increase in direct-to-consumer deliveries will be a challenge to be dealt with. From an ecological standpoint, this increase in demand for last-mile delivery in combination with the sustained growth in ride hailing and ride sharing \cite{yaraghi2017current,hahn2017ridesharing,kokalitcheva2016uber} provides a crucial opportunity for the adoption of more sustainable transportation systems. 

\

The growth in demand for the aforementioned forms of services coupled with the rise of self-driving technology points towards a need for a fleet management framework that combines passenger transportation and goods delivery in an optimal and sustainable manner. To address this, we propose FlexPool -- a distributed reinforcement learning framework to manage a fleet of autonomous vehicles that provide passenger ride-sharing service along with a set of services that include good delivery requests of different service types. Service types include last mile postal delivery, food orders, or grocery delivery to mention a few. This intelligent transportation system which is driven by the objective of maximizing utility and maintaining service levels has the potential to revolutionize transportation systems. 

\subsection{Related Work}

With the shared economy being increasingly in the spotlight, related operational and strategic methods of providing joint transportation services for both people and goods have received research attention. This is coupled with the ever-growing conflict between the increasing demand for mobility and limitations in resources such as fuel. As a result, many researchers have addressed the problem of joint transportation, however the majority rely on an accurate model that represents the environment dynamics. Ours is the first paper to provide a model-free algorithm which is capable of learning and adapting through its interaction with the environment, to the best of our knowledge. 
In this section, we shall discuss the key insights from related works on the joint transportation problem for passengers \& goods as well as the model-free approaches.

\subsubsection{Joint Passengers  \& Goods Transportation}
 Various publications have attempted to address the problem of joint passenger \& goods transportation in varying approaches. Crowddeliver \cite{crowddeliver} is an approach for express package delivery which exploits relays of taxis and passengers to help transport packages collectively without degrading the quality of service for passengers. The authors solve a route planning problem by finding the optimal package operational paths for each package request with the objective of minimizing the package delivery time. This work was one of the first to use a package relaying methodology to improve taxi utilization. However, crowddeliver does not allow for package delivery during passenger rush hours, nor does it assign packages while passengers are being transported (ride-sharing is not considered). In a follow-up paper \cite{CIT}, the authors outline key challenges with joint transportation including (i) the lack of research and data on the spatio-temporal patterns of goods demands where a potential solution is to use interchange stations (referred to as ``hop-zones'' in our paper) as a potential solution. (ii) the need for algorithms that adaptively schedule the crowdsourced resources according to the real-time passenger and \& goods flow information to the stochastic dynamics and uncertainty in a real-time setting; which is a core aspect of our proposed model-free dispatch algorithm. 
 
 Amongst other attempts to develop joint transportation models, the authors of \cite{nguyen2015people} consider problems in which people and parcels are handled in an integrated way by the same taxi network in the city of Tokyo. The authors of \cite{ghilas2013integrating} studied the possibility of transporting freight by public transport. However, these approaches use predetermined routes and schedules. The authors of \cite{masson2017optimization} designed a two-tier distribution system to deliver parcels to shops and administrations located in congested city cores that utilizes the spare capacity of the buses combined with a fleet of near-zero emission city freighters. In \cite{fatnassi2015planning},  the potential of integrating shared goods and passengers was investigated on on-demand rapid transit systems in urban areas. The authors of \cite{chen2018multi} proposed PPtaxi, which is one of the few frameworks solving the joint problem of passenger and goods transportation, with multi-hop driver-parcel matching. They propose an ILP (integer linear programming) formulation for this problem. This is a model-based approach which is not capable of learning and adapting its policy with new observations or data. Further, they do not consider pooling capability for the passengers.

\subsubsection{Model-free Approaches for Transportation}
Model-free approaches have surged in popularity across a variety of fields with the development of Deep Reinforcement Learning \cite{DBLP}. Within the space of intelligent transportation systems and urban planning, models are often used to represent the dynamics of a system environment. With the availability of large dataset \cite{taxi2018limousine} and environments' complex input-output interactions, deep reinforcement learning models provide a means to learn system dynamics using rich function approximators that provide a low dimensional representation the environment \cite{schultz2018deep}.

 Specific to the passenger delivery problem, several studies proved model-free approaches as effective means of learning environment dynamics \cite{oda2018movi,al2019deeppool,singh2019distributed}. MOVI, proposed in \cite{oda2018movi}, addressed the passenger pickup problem for autonomous taxis using a distributed model-free approach for dynamic fleet management. In MOVI, each vehicle solves its own DQN to learn optimal dispatch policies. By training the fleet to minimize an objective defined by demand and supply gap using the New York Taxi trip datasets such as \cite{taxi2018limousine}, the study improves global performance metrics such as passenger accept rate, passenger waiting time, fleet utilization, and fleet fuel cost in comparison to model-based approaches. DeepPool, proposed in \cite{al2019deeppool}, extended MOVI framework to allow ride-sharing of passengers. In DeepPool, vehicles are matched with multiple passengers taking into consideration vehicle seating capacity. DeepPool's distributed DQN objective rewarded vehicles for pooling multiple passengers. This encouraged ride-sharing for improved vehicle utilization. The study proved that extending the distributed model-free approach to the ride-sharing scenario improved passenger accept rate, passenger waiting time, fleet utilization, and fleet fuel cost in comparison to MOVI and other model-based approaches. Consequently, MHRS was introduced in \cite{singh2019distributed} as a multi-hop ride-sharing algorithm where passengers may take multiple transits or ``hops" before their final destination. The action space of this DQN allowed agents to drop off passengers at hop-zones. Their reward was however penalized to prioritize passengers which have already been through hops, to allow for multi-hop ride-sharing where passengers do not have to make an excessive number of stops. MHRS as a result demonstrated an improvement in passenger accept rate, waiting time, fleet utilization, and fuel costs. However, MHRS requires a significant amount of practical incentives for passengers to accept being dropped off at non-terminal locations. Last-mile goods on the other hand are not as sensitive to such inconveniences as long as their delivery is fulfilled in an acceptable time frame. All these approaches demonstrate that model-free approaches are effective, but none of them consider the transportation of goods along with the passengers, thereby underutilizing vehicle carrying capacities such as trunk space, which is the focus of this paper.

\subsection{ Contributions}

The key contributions of the paper are as follows:
\begin{enumerate}[leftmargin=*]
\item This paper provides a distributed algorithm that allows for joint passenger and good transportation, allowing for pooling of passengers in a vehicle, as well as for the goods to transfer vehicles and be transferred in multiple hops (Figure \ref{fig:MHGD} explains the multi-hop scenario for goods). 
\item The key aspects of the proposed algorithm are dispatch of the idle vehicles and the matching of the passengers and goods to the vehicles. The dispatching is performed using a deep reinforcement learning-based approach, where the global objectives are distributed across the vehicles. The distributed approach allows a significant reduction in complexity. The matching is performed for passengers and goods delivery requests considering availability in surrounding vehicles' seating and trunk capacities respectively. Request-to-vehicle assignments are made using a greedy approach to minimize customer waiting time. 
\item The proposed model-free algorithm does not rely on an accurate predefined model of the system. Indeed, it is the first model-free algorithm that considers dispatching and matching for the passengers and goods together. 
\item The proposed algorithm is implemented in a simulator based on New York City taxi-cab data and customer check-in traffic data from Google Maps. Our simulations on joint passenger \& goods transportation demonstrate that FlexPool with Multi-Hop transit for goods improves fleet utilization by 35\% and fuel consumption by 30\% in comparison to (i) model-free approaches which do not consider Multi-Hop transit for goods, and (ii) model-free approaches which do not combine passengers and goods in making assignments to fleet vehicles.
\end{enumerate}

\subsection{Organization}

The rest of the paper is organized as follows. Section II delineates the problem of hybrid workload delivery systems with examples. Section III discusses the framework and describes the proposed algorithm. Section IV presents the simulator used to evaluate the framework. Section V describes the experiments and results that show the performance of the proposed algorithm in comparison to the considered baselines. Finally, Section VI concludes the paper with a discussion of future research. 

\subsection{Abbreviations and Acronyms}\label{AA}
RL: Reinforcement Learning. DQN: Deep Q-Network. DDQN: Double Deep Q-Network. Conv-Net: Convolutional Neural Network.

%% file: statement.tex
\section{System Model}

In this section, we will describe pooling for the passengers, multi-hop transfers for the goods, and the combination of them for the overall FlexPool framework. This will be followed by the key model parameters and notations. 

\subsection{Ride-Sharing  for Passengers}
\begin{figure}
	\begin{center}
		\includegraphics[trim=0.1in 2.1502in 0.2in .0in, clip,width=0.48\textwidth]{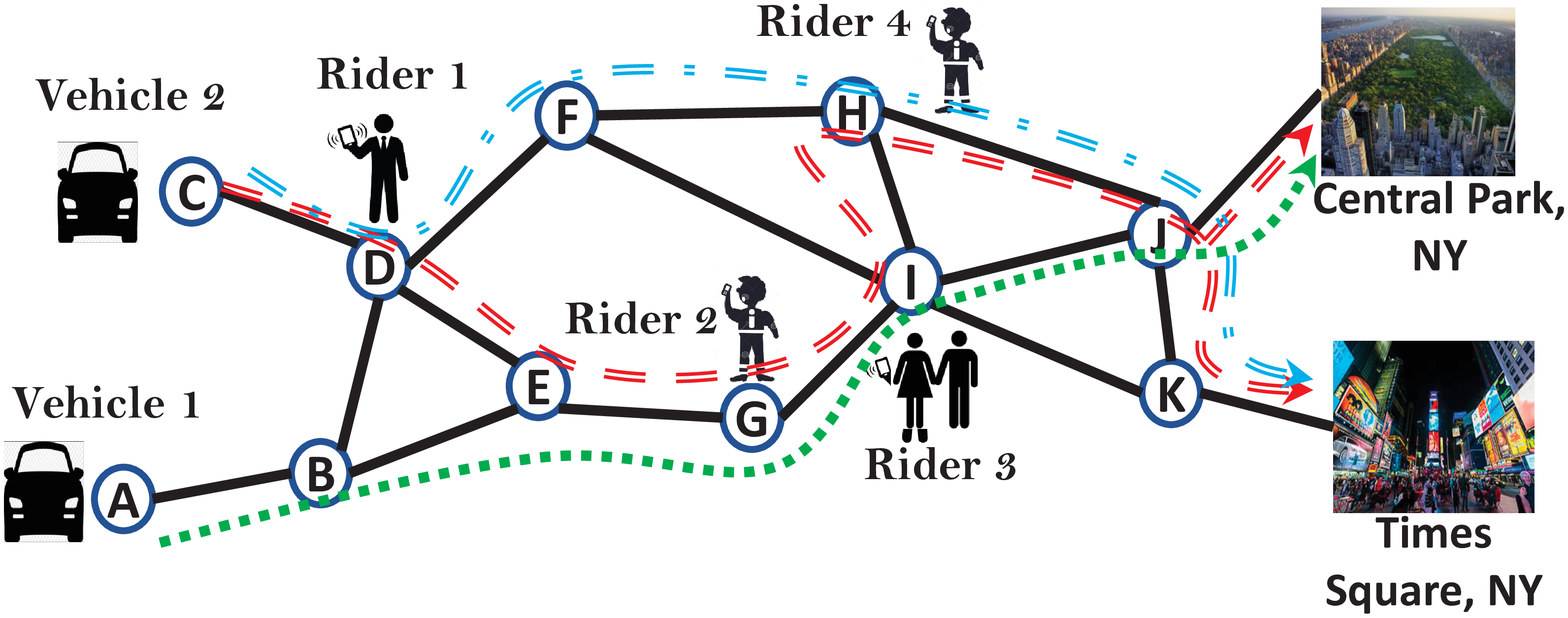}
		\vspace{-.1in}
		\caption{A schematic to illustrate the ride-sharing routing in a region graph, \text{black}{ consisting of 11 regions, $A$ to $K$.} There are four ride requests and two vehicles. The locations of both customers and vehicles are shown in the figure above. Two different possible scenarios to serve the ride requests are shown in the figure and depicted by the dashed-red, dotted-green, and dashed-dotted blue lines. The destination of Rider 1 and Rider 4 is the Central Park, NY, while the destination of Rider 2 and Rider 3 is Times Square, NY.  	
		}
		\label{rideSharing_Example}
		\vspace{-.2in}
	\end{center}
\end{figure} 

Figure \ref{rideSharing_Example} presents an example depicting a real-life scenario. Consider the scenario shown in Figure \ref{rideSharing_Example}, where Rider 1 and Rider 4 want to go to the Times Squares, NY, while
Rider 2, and Rider 3 want to go to the Central Park, NY.
The locations of the four customers are depicted in the figure. Also, there are two vehicles located at node (zone) $A$ and node $C$\footnote{We use node, zone, and region interchangeably. However, a node also can refer to a certain location inside a region/zone.}. Without loss of generality, we assume the capacity of the vehicle ($2$) at location $C$ is $5$ passengers while that at location $A$ ($1$) is limited to $4$ passengers only. Given the two vehicles, there is more than one way to serve the requests. Two different possibilities are shown in the figure, assuming ride-sharing is possible. The two of the many possibilities are: (i) serving all the ride requests using only the vehicle $2$, depicted by the dashed-red line in the figure, (ii) serving Rider 1 and Rider 4 using the vehicle $2$ (see the dashed-dotted blue line), while Rider $2$ and Rider $3$ are served using the vehicle $1$(see the dotted-green line). 

If ride-sharing is not allowed, only two ride requests among the four can get served at a given time. For example, vehicle $2$ needs to pick up and dispatch rider $1$ first, then, it can pick up rider $4$. Though rider $4$'s location is inside the route of the rider $1$s source and destination. Hence, using ride-sharing, fewer resources (only one vehicle, at location $C$, out of the two) are used and consequently less fuel and emission are consumed \cite{shaheen2018shared}. 
Further, the payment cost per rider should be lower because of sharing the cost among all riders. Besides, ride-sharing reduces traffic congestion and vehicle emission by better utilizing the vehicle seats. Thus, ride-sharing will bring benefits to the driver, riders, and society.

\subsection{Multi-Hop Delivery  for Goods}

\begin{figure}
	\includegraphics[width=7.7cm]{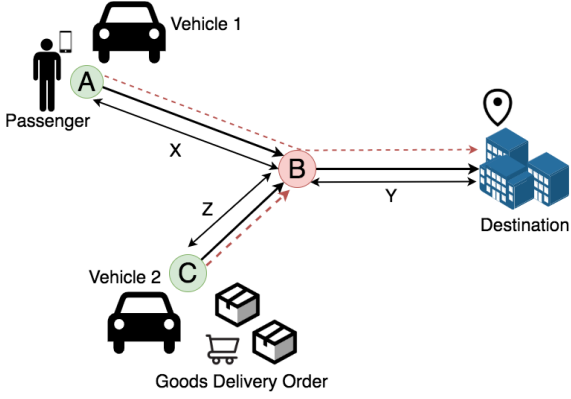}
	\caption{An illustration of multi-hop transportation scenario with hybrid delivery workload. Vehicle 1 and 2 are in zones A and C respectively. Zone B represents a hop-zone' where the packages part of the goods order packages transfer over to another vehicle}
	\label{fig:MHGD}
\end{figure}

In Figure \ref{fig:MHGD}, a good request is to go from zone C to the destination. The good can be transported by one vehicle from C to B, and another vehicle from B to the destination. Zone B will serve as a transit location referred to in this paper as a ``hop-zone''.  This flexibility of changing vehicle is multi-hop transport of goods. Such multi-hop flexibility improves the packing of the goods, as was shown for the case of passengers in \cite{singh2019distributed}. For passenger transfers, it was shown that the multi-hop transfers leads to 30\% lower cost and 20\%
more efficient utilization of fleets, as compared to the ride-sharing
algorithms. Even though such transfers may not be convenient for passengers, they can be used for the goods which motivates the choice in this paper.

\subsection{FlexPool Framework}

In our proposed framework, we consider an environment with both passenger \& goods pick-up requests where vehicles are capable of carrying both types of pick-ups simultaneously. 
We let vehicle capacity be represented by $C = (C_p, C_k)$, where $C_p$ is passenger carrying capacity and $C_k$ is goods carrying capacity. Integrating the multi-hop goods delivery and ride-sharing services can minimize the total distance travelled while improving the number of requests (passengers \& goods both) by the vehicles. Packages can be picked up and dropped off from hop-zones, while passengers will be seated from the point they are picked up till the point they are dropped off. Given such a scenario, assignments will be done to vehicles in a first come first serve manner regardless of package or passenger under the defined capacity constraints.

To illustrate the FlexPool framework, we consider one package and one goods delivery request as depicted in  Figure \ref{fig:MHGD}. Both the passenger \& goods order (set of packages) are to be dropped off at the same destination, while are requested from zones A and C, respectively.  Given zone B lies along the delivery route of both requests, Zone B can serve as a transit location. In this scenario, one vehicle (say Vehicle 2) can drop its packages to consequently be picked by another vehicle (Vehicle 1) to deliver to the final destination. 
Accordingly, the passenger initially gets picked up by the nearest vehicle (Vehicle 1) and the packages by Vehicle 2. Vehicle 2 drops the packages at hop zone B. As Vehicle 1 reaches the hop-zone (Zone B), it picks up the packages and drops both passengers and packages at their destination. As shown in the figure \ref{fig:MHGD}, X represents the distance between Zone C and Zone B, Y represents the distance between Zone B and the destination, and Z is the distance between Zone A and Zone C. We define the effective distance traveled by Vehicle 1 ($D_1$)  by the following equation:

\begin{equation}
D_{1} = \frac{X+Y+Y}{X+Y} = 1 + \frac{Y}{X+Y}
\end{equation}

Effective distance is formally defined as the ratio of total distance covered if no hoping and sharing was allowed to the total distance covered when hoping and sharing is allowed.  The efficient packing of vehicles decreases the overall distance traveled by the vehicles in completing service of the same number of requests. The multi-hop scenario shown in Figure \ref{fig:MHGD} reduces the traffic that goes from Zone B to the final destination. Additionally, the accept rate of pick-up requests can be improved as vehicles get assigned more frequently. As in our example Vehicle 1 is free after dropping Package 1 and can pick-up new orders accordingly. We note that  the ride-sharing of passengers, multi-hop ride-sharing of goods,  as well as joint packing of passengers and goods, decrease the effective distance of vehicles.

\subsection{Model Parameters and Notations}

In this section, we will describe the notations used to represent, the state, action, and reward spaces. The optimization of the system is achieved over $T$ time steps with each step of length $\Delta t$. The fleet make decisions on where on the map to go to serve at each time step $\tau = t_{0},t_{0}+\Delta t,t_{0}+2(\Delta t),\ldots,t_{0}+T(\Delta t)$ where $t_{0}$ is the start time.

The map is split up into a grid with each square being taken as a  zone. Zones are represented by $i \in  \{1,2,3,\ldots,M\}$.  The number of vehicles in the fleet is represented by $N$. A vehicle is marked as \textit{available} if there is remaining seating or trunk capacity.  Vehicles that are completely full or are not considering taking passengers or goods are marked \textit{unavailable}. Available vehicles in zone $i$ at time slot $t$ is denoted $v_{t,i}$. Only available vehicles are eligible to be dispatched. 

 \textbf{$X_t$}  tracks the vehicle seating capacity and trunk space for packages, denoted as \textbf{$C_{p,v}$} and \textbf{$C_{k,v}$}, respectively. As a result \textbf{$X_t$} will track: 
\begin{enumerate*}
	\item current zone for vehicle $v$,
	\item available seats,
	\item available trunk space, 
	\item time at which delivery order is picked up, and 
	\item destination zone of each delivery order.
\end{enumerate*} 

Pick-up requests at a given zone $i$ are denoted by $\delta_{t,i}$ at time slot $t$. This represents the demand at a given area at that time. 	At each time slot $t$, the supply of vehicles for each zone is projected to future time  $\tilde{t}$. Consequently, the number of vehicles that will become available at  $\tilde{t}$ is denoted by $\delta_{t,\tilde{t},i}$. This value is ascertained from an ETA (estimated time of arrival) prediction for all moving vehicles. Consequently, given a set of dispatch actions, we are able to predict the number of vehicles in each zone for $T$ time slots ahead, denoted by $V_{t:T}$.  The pick-up request demand in each zone is predicted through a historical distribution of trips across the zones, and is denoted by $D_{t:T} = (\overline{d}_{t},\ldots,\overline{d}_{t+T})$ from time $t$ to $t + T$. All the data is combined to represent the environment state space $s_{t}$ by the tuple $(X_{t}, V_{t:T}, D_{t:T})$.
At each assignment of a request to vehicle, the state space tuple is updated with the expected pick-up time, source, and destination data.

%% file: framework.tex
\section{FlexPool Framework and Proposed Algorithm}
This section presents our distributed framework and the proposed algorithm to train each vehicle in the fleet, which we refer to as an agent, to learn the optimal policy. We will first describe the objectives, and then describe the dispatch and matching policies.   The dispatch policy sends unassigned or idle vehicles to  regions of anticipated future demand in order to pick up future requests.  The matching policy assigns vehicles to pickup nearby requests after the vehicles are dispatched.  These will then be combined to provide the overall algorithm.

\subsection{Objectives for the Proposed Problem}

The goals of the Flexpool framework include: (i) satisfying the demand of pick-up orders, thereby minimizing the demand-supply mismatch, (ii) minimizing the time taken to pick-up an order (aka pick-up wait time) in tandem to the dispatch time taken for a vehicle to move to a pick-up location, (iii) minimizing the additional travel time incurred by orders due to participating in a shared vehicle, (iv) minimizing the additional travel time incurred by orders due to layovers at a hop-zone, and (v) minimizing the number of vehicles deployed to minimize fuel consumption and traffic congestion while maximizing utility of available capacity within the available vehicles. These five objectives are studied through the five components, described below.  The first component aims to minimize the gap between the supply and demand of order pick-ups. Equation (\ref{eq:reward-1}) represents this with $v_{t,i}$ is the number of vehicles at time $t$ in zone $i$. Hence we have the supply demand difference accounted for each time slot $t$ at each zone $i$ given by $(\bar{d}_{t,i} - v_{t,i})$
\begin{equation}
\text{diff}_{t}^{(D)} = \sum_{i=1}^{M} (\bar{d}_{t,i} - v_{t,i})^{+}
\label{eq:reward-1}
\end{equation}
where $(\cdot)^+ = \max (0,\cdot)$.

The second component aims to minimize the dispatch time for each vehicle, i.e., time taken for a vehicle to travel current zone from to a zone where pick-ups are made. Available vehicles may be dispatched in two cases: (i) Serve a new request, or (ii) move to locations where a future demand is anticipated. In equation (\ref{eq:reward-2}), $h_{t,j}^{n}$ represents the estimated travel time for vehicle $n$ to arrive at zone $j$ at time $t$. For all vehicles available within time $t$, we seek to minimize the total dispatch time $T_{t}^{(D)}$, 
\begin{equation}
T_{t}^{(D)} = \sum_{n=1}^{N} \sum_{j=1}^{M} h_{t,j}^{n} u_{t,j}^{n} 
\label{eq:reward-2}
\end{equation}
where $u_{t,j}^{n} = 1$ only if vehicle $n$ is dispatched to zone $j$ at time $t$ and $0$ if otherwise.
Minimizing dispatch time ultimately minimizes fuel costs when a vehicle is gaining revenue by serving customers.

The third component minimizes the extra travel time for every order (both passengers and packages) that is incurred due to vehicle sharing. For each vehicle $n$ which is carrying $l$ delivery orders of any given kind at each time step $t$, we minimize $\delta_{t,n,l} = t^{'} + t_{t,n,l}^{\left(a\right)} - t_{n,l)}^{\left(m\right)}$, where
$t_{t,n,l}^{\left(a\right)}$ is the updated time the vehicle will take to drop-off passenger/package $l$ due to a change in route and/or addition of another order at time $t$.  

$t_{n,l}^{(m)}$ is the travel time that would have been taken if the picked up order $l$ would have travelled without sharing the vehicle with other orders. $t^{'}$ is the time elapsed after order $l$ was put in the request queue. Equation \ref{eq:reward-3} takes all the above into consideration.
The following component expresses, $\Delta_{t}$, the total extra travel overhead time:
\begin{equation}
\Delta_{t} = \sum_{n=1}^{N}\sum_{l=1}^{U_{n}} \delta_{t,n,l}
\label{eq:reward-3}
\end{equation}
Note that a given vehicle $n$ may not know the destination of the picked up order. As a results, it will take the expected time of order's travel generated in any region. Hence, $\delta_{t,n,l}$ is assumed to be the mean value.
To minimize the inconvenience of transferring packages at hop-zones to passengers currently being served in a vehicle as well as potential delay in delivery of packages, we introduced the following component which minimizes the number of hops packages are made to go through before being a delivery is fulfilled. Here we define $g$ as the number of packages being carried by a vehicle. $\text{hop}_{g,i,p}$ denotes that package $g$ has undergone a hop-transfer at zone $i$, and $p$ denotes the $p^{\text{th}}$ time the package is being transferred along its journey which is defines as follows:
\begin{equation}
H_{t} = \sum_{i=1}^{M}\sum_{p=1}^{P} \text{hop}_{g,i,p}
\label{eq:reward-4}
\end{equation}
Equation (\ref{eq:reward-4}) provides a sum total of all hop-transfers of all packages being delivered at time step $t$. 

The final component aims to minimize the number of vehicles in the fleet. To achieve this, we minimize $e_t$ which is indicative of the total vehicles being used at time step $t$. The number of vehicles being used at tine $t$ is given as follows:

\begin{equation}
e_{t} = \sum_{n=1}^{N} \max(e_{t,n} - e_{t-1,n}, 0)
\label{eq:reward-5}
\end{equation}
In equation (\ref{eq:reward-5}), $e_{t,n}$ indicates whether a vehicle is active. By minimizing the number of fleet vehicle deployed on the roads, we achieve a better utilization per resource. This also results in a reduction of idle cruising time which in turn mitigates extraneous fuel consumption.

The overall objective is a linear combination of all the above described components as follows: 
\begin{equation}
\overline{r} = -[\beta_1 \text{diff}_{t}^{(D)} + \beta_2 T_{t}^{(D)} + \beta_3 \Delta_{t} + \beta_4 e_{t} + \beta_5 H_{t}],
\label{eq:reward-6}
\end{equation}

where the minus sign in the above indicates we want to minimize these terms.

\subsection{Distributed Dispatch Framework}

In this subsection, we describe the distributed framework for dispatch of idle vehicles. For the dispatch, we utilize a reinforcement learning framework, with which we can learn the probabilistic dependence between vehicle actions and the reward function thereby optimizing our objective function as in \cite{oda2018movi,al2019deeppool,singh2019distributed}. In the following, we describe the state, action, and reward for the dispatch policy.

\textbf{\textit{State}}: The state variables defined in this framework capture the environment status and thus influence the reward feedback to the agents' actions. We discretize the map of our urban area into a grid of length $\boldsymbol{x}$ and height $\boldsymbol{y}$; resulting a total of $\boldsymbol{x}*\boldsymbol{y}$ zones. This discretization prevents our state and action space from exploding thereby making implementation feasible. The state at time $t$ is captured by following tuple: $(X_{t},V_{t:T} , D_{t:T})$. These elements are combined and represented in one vector denoted as $s_{t}$. When a set of new ride requests are generated, the FlexPool engine updates its own data to keep track the environment status. The three-tuple state variables in $s_{t}$ are passed as an input to the DQN input layer which consequently outputs the best action to be taken.
\begin{enumerate}
	\item 
	\textbf{$X_t$} will track vehicle seating capacity and trunk space for goods/packages. \textbf{$C_{p,v}$} and \textbf{$C_{k,v}$} respectively. As a result \textbf{$X_t$} will track: \textit{current zone of vehicle $v$, available seats, available trunk space, pick-up time of delivery order, destination zone of each order}. 
	\item
	\textbf{$V_{t:T}$} is a prediction of number of available vehicles at each zone for $T$ time slots ahead. 
	\item
	\textbf{$D_{t:T}$} has a term \textbf{$\delta_{t:T}$} that predicts joint demand of passengers \& goods delivery orders at each zone for $T$ time slots ahead. 
\end{enumerate}

\textbf{\textit{Action}}: The action of vehicle $n$ is denoted by  $a_{t,n}$ which consists of two components:
\begin{enumerate*}
	\item  if the vehicle is partially filled, it decides whether to serve the existing customers (passengers or goods) or to accept new requests, and
	\item if it decides to serve a new request or the vehicle is totally empty, it decides the zone it should be dispatched to at time slot t.
\end{enumerate*} 
If vehicle $n$ is full it can not serve any additional customer. Alternatively, if a vehicle decides to serve current customers, the shortest route is used along the road network to pickup the assigned orders.

\textbf{\textit{Reward}}: 
Below is the reward function $r_{t,n} = r(s_{t,n} , a_{t,n})$ which is used at each agent $n$ of the distributed system at time $t$. The weights shown in equation (\ref{eq:DQNreward}) are used only in instances when an agent is not fully occupied and is eligible to pickup additional orders. 
\begin{equation}
\begin{split}
r_{t,n} &= \beta_1(b_{t,n} + p_{t,n}) - \beta_2c_{t,n} - \beta_3 \sum_{u=1}^{U_n} \omega_u \delta_{t,n,u} \\ 
&\text{ } - \beta_4 \max (e_{t,n} - e_{t-1,n}, 0) - \beta_5 \max_{u\in\mathcal{L}_{t,n}}H_u
\end{split}
\label{eq:DQNreward}
\end{equation}

Here, $b_{t,n}$ denotes the number of passengers served by vehicle $n$ at time $t$ and $p_{t,n}$ denotes the number of packages being carried in the trunk of vehicle $n$ at time $t$. The $\beta_1$ term rewards agents for picking up more requests to meet demands.\newline

 $c_{t,n}$ denotes the time taken by vehicle $n$ at time $t$ to hop or take detours to pick up extra orders. This term discourages the agent from picking up additional orders without considering the delay in current passengers/goods orders. As a result, the $\beta_2$ term prevents adverse effects on customer travel time due to ridesharing. \newline

Further, $\sum_{u=1}^{U_n} \omega_u \cdot \delta_{t,n,u}$ denotes the sum of additional time vehicle $n$ is incurring at time $t$ to serve additional passengers or packages. The $\omega_u$ term is an ``urgency" weight for the $u$'th order. This is an attribute assigned based on the type of request (passenger or good) being transported. Ride-sharing passenger requests may be assigned $\omega = 1$, whereas goods may be assigned $\omega < 1$ since their travel is not as affected by urgency or convenience. Overall, this component penalizes the agent for decisions that delay the transportation of passengers. As a result, agents are incentivized through the $\beta_4$ term to prevent loss of customer convenience. \newline

$\max (e_{t,n} - e_{t-1,n}, 0)$ addresses the objective of minimizing the number of vehicles at $t$ to improve vehicle utilization. Minimizing the $\beta_4$ term reduces overall fuel consumption of the fleet. \newline

$ \max_{u\in\mathcal{L}_{t,n}}H_u$ is the max of the number of hops done by packages in a given vehicle. The $\beta_5$ term as a result incentivizing agents to drop-off packages that have been through hop-zones to their final destination, thereby minimizing the number of hops taken by packages.

 This reward function is a distributed version of the global objective in \eqref{eq:reward-6}. With reinforcement learning, we build a representation of the environment at each time step $t$ by a state $s_t$ and reward $r_t$. Using this information, an action $a_t$ is chosen to direct (dispatch) available vehicles to different locations such that the expected discounted future reward, $\sum_{k=t}^{\infty} \eta^{k-t} r(a_{t},s_{t})$, is maximized, where $\eta< 1$ is a discount factor.  

\input{hopzone_algo.tex}

\subsection{Hop-zone Designation}

	As previously mentioned, a hop-zone is a location where a package is dropped off in transit during its journey to its destination. 
	In our algorithm, only goods are assigned hop-zones. Hop-zones are pre-determined locations on the map and are assumed to have the necessary storage infrastructure to hold a large number of packages. As proposed in \cite{crowddeliver} and \cite{CIT}, POI locations such as gas stations and convenience stores may be incentivized to offer storage services to the transportation system to enable such an infrastructure. 
	 
	We use a heuristic approach to assign hop-zones to goods delivery requests as shown in Algorithm \ref{alg:hop}. For each request, we find the nearest hop-zone such that the total delivery distance is less than two times the original distance. If such a hop-zone exists, the trip is split up recursively into hop-trips until no suitable hop-zone assignment can be made upon which the package is delivered to its final destination.

\subsection{Matching Algorithm}
Given the hop-zone locations decided, and the dispatch algorithm using DDQN, we now describe the matching algorithm, detailed in Algorithm \ref{alg:matching}, for the passengers and goods to each vehicle. As seen in lines 1 and 2 of Algorithm \ref{alg:matching}, the inputs for the matching algorithm are vehicle status variable \textbf{$X_t$} and the set of requests within a given time-step. The matching algorithm matches the pick-up requests to the vehicle in a greedy fashion minimizing the waiting time of passengers or goods. As seen in lines 3 to 5 of Algorithm \ref{alg:matching}, we consider all vehicles present within the region where a request is submitted and calculate the ETA, available seating capacity, and available trunk capacity. In Lines 7-10 of Algorithm \ref{alg:matching}, if the request is a passenger, it is assigned to the nearest vehicle with vacant seats. Whereas if the request is a goods package, it is assigned to the nearest vehicle with vacant trunk space. We note that the matching for the goods and passengers are done in parallel, and if a passenger/good is assigned to multiple vehicles, one is chosen uniformly at random. This maintains the distributed nature of the algorithm. As a result, greedy allocation of passengers and packages are done until either there are no more requests to be assigned or all available vehicles have been fully occupied.

\input{matching_algo.tex}

\input{dispatch_algo.tex}

\subsection{Double Deep Q-Learning (DDQN)}
For each vehicle $n$, a dispatch decision is taken when the vehicle is idle. This policy is learned from the Double Deep Q-Networks (DDQN) approach described in Algorithm \ref{DoubleDQN_alg}. The output of the DDQN is the Q-values corresponding to a discrete set of dispatch actions, while the input is the environment status governed by the vector $\boldsymbol{\varOmega}_{t}$. $\boldsymbol{\varOmega}_{t}$ is defined as the state at any time $t$. Every agent $n$ selects the action that maximizes its reward, i.e., taking the {\it argmax} of the DDQN-network output (line 14 of Algorithm \ref{DoubleDQN_alg}). The learning starts with zero knowledge and actions are chosen following the epsilon-greedy policy where the agent chooses the action that results in the highest Q-value with probability $1 - \epsilon$. Consequently, it selects a random action and gathers more information through exploration at a probability of $\epsilon$. The $\epsilon$ is annealed linearly from 1 to 0.05 over $T_n$ steps. This allows the agent to balance exploration with exploitation for rewards. Additionally, we use experience replay to overcome the issue of instability due to nonlinear approximations from the neural-network, and due to correlations between the action-value. Experience replay involves accumulating a replay memory buffer of many episodes which is input to the DDQN. DDQN implements Q-Learning with two neural-network-based value functions. One is the target network which learns during the experience-replay while the other is the Q-network which is copy of the last episode of the target network. These two networks are represented by weights $\theta$ and $\theta^{'}$ respectively. DDQNs have been proven to be effective means of mitigating Q-value overestimations in \cite{DDQN_arxiv}.
In the DDQN, having two function approximators allows to decouple the process of taking actions and estimating the Q-value. For each update in the DDQN, one set of weights is used to determine the greedy policy and the other to determine its value. The target Q-value at any time step $t$ can be defined as:
\begin{equation}
\boldsymbol{T}_{t}= {R}_{t+1} + \gamma Q({S}_{t+1},{argmax}_{a}Q({S}_{t+1},a;\theta_{t}); {\theta}^{'}_{t})
\end{equation}
\\
where $\theta_{t}$ is the target network used to determine the greedy policy and ${\theta}^{'}_{t}$ is the Q-network used to evaluate the value of this policy. 

\input{flexpool_algo.tex}

\subsection{FlexPool Algorithm}

The overall algorithm is shown as Algorithm \ref{alg:Algo}. The inputs are a state vector $\Omega_{t,n}$, pick-up requests, and map-based locations of vehicles and pick-up requests (line 1 of Algorithm \ref{alg:Algo}). The state vector $\Omega_{t,n}$ is determined by the available vehicles and pickup records which are generated from the passenger-goods dataset described in section IV (line 3 of Algorithm \ref{alg:Algo}). Then, we initialize the number of vehicles and generate some pickup requests in each step based on the dataset (see lines 6-7 of Algorithm \ref{alg:Algo}).   Each vehicle can be in five states - ``Dispatching", ``Serving", ``Matched",  ``Idle", ``Dispatched".  The detailed procedure in each of the states is shown in Algorithm \ref{alg:Algo}. To understand the steps, assume that the vehicle just got empty, in which case it will be marked as ``Idle" (line 14). In the idle state, the dispatch decision from Q-network is used, and the vehicle is dispatched to the new location. The status is updated to ``Dispatching" (line 18-22). In the ``Dispatching"  state, the vehicle goes to the dispatch location, and upon reaching the desired location, the status is updated to ``Dispatched" (line 9-11). If the vehicle has been dispatched, a  matching assignment is done for both passenger and goods requests, which will be based on  Algorithm \ref{alg:hop} and Algorithm \ref{alg:matching} (line 23-37). The vehicle status us set to ``Matched," and the vehicle drives to pickup locations and on route to the patch of pickup and delivery of assigned passengers and goods. The extra travel time $\delta_{t,n}$ and hop counter $H_{t,n}$ are updated as needed (see lines 35-36 of Algorithm \ref{alg:Algo}).  If the vehicle has been matched and has crossed the pickup location, it is marked as ``Serving" (line 15-17). After all the passengers and goods that were matched have been delivered, the vehicle status changes to ``Idle" (line 12-14). The procedure repeats itself going through these stages.

As mentioned previously, the output of the DDQN is the Q-values for each movement possible on the map for a given vehicle. Note that each update is performed in parallel across all agents. Each agent however does not anticipate the actions of other agents on the map thus limiting coordination amongst them. Consequently, the vehicles travel to their chosen dispatch locations by using the shortest path on the road network graph as seen in line 20 of Algorithm \ref{alg:Algo}. It is to be noted that FlexPool can is scalable to a large number of vehicles and requests. The factors that enable these are:  \begin{enumerate*}  
	\item each agent solves its own DDQN which runs at a time complexity of O(1) during inference, and 
	\item the distributed nature of our algorithm allows each vehicle to take a discrete set of actions. This prevents explosion in action space since we do not consider the joint action space across all agents.
\end{enumerate*} 

%% file: hopzone_algo.tex
\begin{algorithm}
	\caption{Assign Hop-zone }\label{alg:hop}
	\begin{algorithmic}[1]
		\State \textbf{Inputs}: Initial Request (Origin, Destination)
		\State \textbf{Outputs}: Hop-trips
		\State \textbf{Initialize} Hop-trips as a set: [(Origin, Destination)]
		
		\For {trip in Hop-trips}
			\State \textbf{Compute} original delivery distance (Origin to Destination)
			\State \textbf{Compute} distances from Origin to Hop-zones
			\State \textbf{Assign} Hop-zone as the nearest Hop-zone
			\State \textbf{Compute} total delivery distance (Origin to nearest hop-zone to Destination)
			\If {total delivery distance $> 2 *$original delivery distance}
				\State \textbf{Return} Hop-trips
			\Else
				\State \textbf{Update} Hop-trips as [(Origin, Hop-zone), (Hop-zone, Destination)]
				\For {trip in Hop-trips}
					\State \textbf{Update} Hop-trips using algorithm \ref{alg:hop}
				\EndFor
			\EndIf
		\EndFor
		\State \textbf{Return} Hop-trips
	\end{algorithmic}

\end{algorithm}

%% file: matching_algo.tex
\begin{algorithm}
	\caption{Matching Algorithm}\label{alg:matching}
	\begin{algorithmic}[1]
		\State \textbf{Inputs}: Demand $D_{t}$,Vehicle State Variable $x_{t,n}$, 
		\State \textbf{Outputs}: Vehicle to Request Assignments
		\If {no available requests within reject radius}
		\State \textbf{Return} None
		\EndIf
		\State \textbf{Sort} requests in ascending order by ETA to pickup location 
		\For {each request within reject radius of pickup}
			\If {request is passenger}
				\State \textbf{Calculate} remaining seating capacity $C_{p,n}$
				\If {$C_{p,n} > 0$}
					\State \textbf{Assign} passenger to seat
				\Else
					\State \textbf{Continue}
				\EndIf
			\ElsIf {request is goods}
				\State \textbf{Calculate} remaining trunk capacity $C_{k,n}$
				\If {$C_{k,n} > 0$}
					\State \textbf{Assign} package to trunk
				\Else
					\State \textbf{Continue}
				\EndIf
			\EndIf
		\EndFor
	\State \textbf{Return} Vehicle to Request Assignments
	\end{algorithmic}

\end{algorithm}

%% file: dispatch_algo.tex
\begin{algorithm}[t]
	\caption{ Double Deep Q-learning with experience replay}
	\label{DoubleDQN_alg}
	\begin{algorithmic}[1]
		\State \textbf{Initialize} replay memory $D$, Q-network parameter $\theta$, and target Q-network $\theta^-$.
		\For{$e:1:Episodes$}
		\State \textbf{Initialize} the simulation with arbitrary number of vehicles and ride requests based on real data.
		\For{$t:\Delta t:T$}
		\State \textbf{Perform} the dispatch and match order
		\State \textbf{Update} the state vector $\varOmega_{t}=$ $(X_t,  V_{t:T},  D_{t:T})$.
		\State \textbf{Update} the reward $r_t$ based on actions $a_t$.
		\For {all available vehicles $n$}
		\State \textbf{Create} $\varOmega_{t,n}=$ $(X_{t,n},  V_{t:T},  D_{t:T})$.
		\State \textbf{Store} transition
		\State $(\varOmega_{t-1,n}, a_{t-1,n},r_{t,n},\varOmega_{t,n},c_{t,n})$
		\EndFor	
		\State \textbf{Sample} random transitions 
		\State $(\varOmega_i, a_i,r_i,\varOmega_{i+1},c_{i+1})$ from $D$.
		\State Set $a_{i}^{*}=\text{argmax}_{a^{'}}Q(\varOmega_{i+1},a_{i}^{*};\theta^{-})$.
		\State Set $z_{i}=r_{i}+\gamma^{1+c_{i+1}}\hat{Q}(\varOmega_{i+1},a_{i}^{*};\theta^{-})$.
		\State \textbf{minimize} $(z_{i}-Q(\varOmega_{i},a_{i};\theta))$ w.r.t. $\theta$.
		\State Set $\theta=\theta^-$ every $N$ steps.
		\State \textbf{Update} the set of available vehicles $A_{t}$
		\For {$n$ in $A_t$}
		\State \textbf{Create} $\varOmega_{t,n}=$ $(X_{t,n},  V_{t:T},  D_{t:T})$.
		\State \textbf{Choose}, with prob. $\epsilon$,  a random action from 
		\State $a_t^{(n)}$.
		\State \textbf{Else} set $a_{t}^{(n)}=\text{argmax}_{a}Q(\varOmega_{t}^{(n)},a;\theta)$.
		\State \textbf{Send} vehicle $n$ to its destination, based on 
		\State $a_{t}^{(n)}$.
		\State \textbf{Update}  $\varOmega_{t,n}$.
		\EndFor
		\EndFor
		\EndFor  
	\end{algorithmic}
\end{algorithm}

%% file: flexpool_algo.tex
	\begin{algorithm}
	\caption{FlexPool Algorithm}\label{alg:Algo}
	\begin{algorithmic}[1]
			\State \textbf{Inputs}: Environment State Vector $\Omega_t$, Pick-up Requests, Map-based locations.
		\State \textbf{Outputs}: Decisions for Matching and Dispatching.
		\State \textbf{Create} state vector $\Omega_{t,n} = (X_{t}, V_{t:t+T}, D_{t:t+T})$
		\State \textbf{Initialize} vehicle states $X_0$ as location of first N requests. 
		\For {$t \in T$}
			\State \textbf{Fetch} all pickup requests at time slot t, $D_{t}$
			\State \textbf{Fetch} all available vehicles at time slot t, $V_{t}$
			\For {each vehicle $V_n \in V_t$}
				\If {$x_{t,n}^{status}$ is \textbf{``Dispatching''}}
					\If {$x_{t,n}^{location} == a_{n}$}
						\State \textbf{Update} $x_{t,n}^{status}$ to \textbf{``Dispatched''}
					\EndIf
				\ElsIf {$x_{t,n}^{status}$ is \textbf{``Serving''}}
					\If {all requests delivered}:
						\State \textbf{Update} $x_{t,n}^{status}$ to \textbf{``Idle''}
					\EndIf
				\ElsIf {$x_{t,n}^{status}$ is \textbf{``Matched''}}
					\If {Vehicle arrived at pickup location}:
						\State \textbf{Update} $x_{t,n}^{status}$ to \textbf{``Serving''}
					\EndIf
				\EndIf
				\If {$x_{t,n}^{status}$ is \textbf{``Idle"}}
					\State \textbf{Push} the state vector $\Omega_{t,n}$ to agent
					\State \textbf{Get} the optimal dispatch action $a_{t,n} $ from Q-network
					\State \textbf{Update} $x_{t,n}^{status}$ to \textbf{``Dispatching''}
					\State \textbf{Drive} to dipatch location $a_t$ taking the shortest path
					\EndIf
				\If {$x_{t,n}^{status}$ is \textbf{``Dispatched"}}
					\For {each passenger request $\in D_t$}
						\State \textbf{Get} passenger matching using Algorithm \ref{alg:matching}
						\State \textbf{Update} vehicle seating capacity $C_{p,n}$
					\EndFor
					\For {each goods request $\in D_t$}
						\State \textbf{Push} goods request to Algorithm \ref{alg:hop} 
						\State \textbf{Get} hop-zone assignments and hop-trips 
						\State \textbf{Update} $D_{t:t+T}$ with hop-trips
						\State \textbf{Get} goods matching using Algorithm \ref{alg:matching}
						\State \textbf{Update} vehicle trunk capacity $C_{k,n}$
					\EndFor
					\State \textbf{Update} $x_{t,n}^{status}$ to \textbf{``Matched''}
					\State \textbf{Estimate} the dispatch and travel time using ETA model
					\State \textbf{Update} $\delta_{t,n}$ if needed, and generate vehicle trajectory
					\State \textbf{Update} $H_{t,n}$ if needed
					\State \textbf{Drive} to pickup requests and update location
				\EndIf
			\EndFor
		\State \textbf{Update} the state vector $\Omega_{t+1}$
		\EndFor
	\end{algorithmic}
\end{algorithm}

%% file: simulator.tex
\section{Simulator Setup}
The fleet of autonomous vehicles were trained in a virtual spatio-temporal environment that simulates urban traffic and routing. In our simulator, we used the road network of the New York City Metropolitan area along with a realistic simulation of taxi pick-ups and package delivery requests. This simulator hosts each deep reinforcement learning agent which acts as a delivery vehicle in the New York City area that is looking to maximize its reward.

\subsection{Dataset}
The delivery workload in this simulator consists of passenger pick-up requests as well as package delivery requests for various goods and services. 
To emulate a realistic workload of passenger pick-ups in the urban environment, we used the New York City taxi trip data set from \cite{taxi2018limousine}.We extracted trips within the major burrows of the metropolitan area from May and June for our simulation.

\begin{figure}
	\includegraphics[width=.47\textwidth]{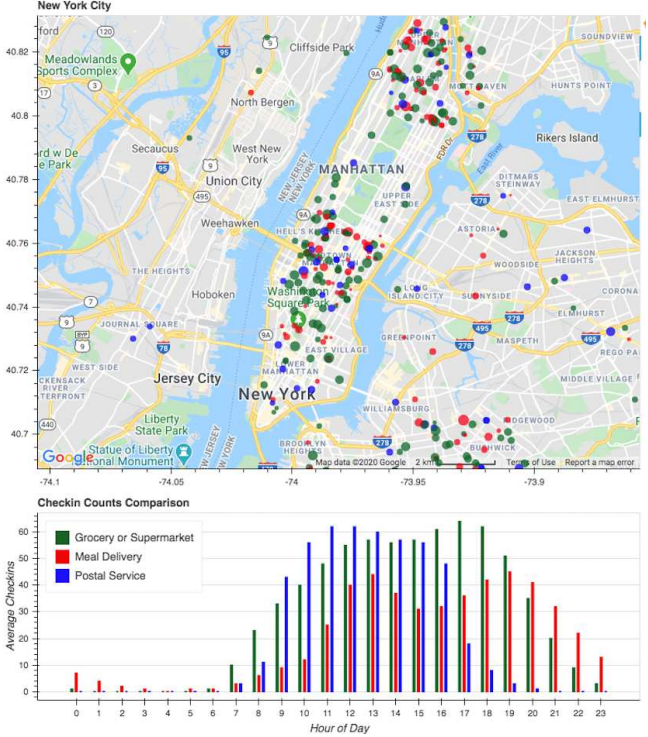}
	\caption{The map and graph show different types of goods delivery requests.}
	\label{fig:Checkins}
\end{figure}

In simulating package delivery requests, customer check-in traffic was extracted from Google Maps for postal service, meal delivery, and supermarket locations. The 100 most active locations were considered from each of the respective service types. Average check-in traffic was extracted for each day of the week for each location. This was used to generate a synthetic workload for a total of two months representative of May and June. Figure \ref{fig:Checkins} shows the distribution of check-ins across the city over a synthetic workload, which was generated from customer check-in from Google Maps analytics data. At each service location, the request rates were generated by Poisson distribution given in equation (\ref{Poisson}) across request rate $x$,  where $\lambda$ represents the observed check-in rate from Google Maps. Consequently, for each service location, package drop-off locations were generated randomly considering delivery radius limit of 5 miles in accordance with the current standard for major crowd-sourced delivery services such as DoorDash and GrubHub. All pick-up and drop-off locations are constrained within the New York City borough boundaries. The resulting data set consisted of goods delivery orders over one month. 

\begin{equation}
p(x; \lambda) = \frac{e^{-\lambda} \lambda^x}{x!}\\
; x \in \mathbb{Z}
\label{Poisson}
\end{equation}
We let the vehicle carrying capacities be $C_p = 4$ passengers and $C_k = 5$ packages, unless stated otherwise. A request is deemed rejected if there are no vehicles available within a radius of $5km^2$. When an adequate number of vehicles are not present to meet the demand of pick-up requests, a higher reject rate can be observed. 

\subsection{Initialization}
A directed graph was constructed as the NYC road network from OpenStreetMaps by partitioning the city into a 212 x 219 service area grid of size 150m x 150m each. In performing routing, pick-up and drop-off locations are snipped to the closest edge-nodes of the network and a shortest path algorithm is found to estimate the travel times. To estimate the minimal travel time for each dispatch, the travel time between every two dispatch nodes/location is ascertained.
 Every third intersection of zones in the horizontal and vertical direction is considered a hop-zone. To allow hop-zones to be adequate transit locations for the majority of the rides, we ensure that the hop-zones are in the busy area of the city and are not too closely placed. As a result, there were a total of 195 hop-zone candidates. To further eliminate under-utilized hop-zones, we consider only the zones with at least 10 pick-up requests in the day. Consequently, we obtain a total of 148 hop-zones in the urban simulation.
 
To initialize the environment, we run the simulation for 100 iterations without dispatching vehicles. At default setting, the simulator initializes 8000 vehicles unless specified otherwise. The maximum horizon is defined as $T=30$ steps where $\Delta T = 1$minute. Unless otherwise stated, the reward function parameters are set to be the following: $\beta_1 = 10, \beta_2 = 1, \beta_3 = 1, \beta_4 = 0.05$,  and $\beta_5 = 2$. We also assign $\omega = 1$ for passengers, and $\omega = 0.5$ for goods to emphasize the minimizing passenger delays. The algorithm also needs estimate of Estimated Time of Arrival (ETA)  and Demand prediction. These estimates are based on deep learning framework and follow the procedure as in \cite{al2019deeppool}. The details are also provided in Appendix \ref{eta_pred} for completeness. 





\subsection{DDQN Implementation}

Each agent is trained using a Double Deep Q-Network (DDQN) algorithm. The dispatch action space of a given agent is limited to 7 grid boxes vertically and horizontally.  To achieve this, the New York City area map was divided into $41 \times 43$ grids. As a result, each vehicle is able to make a decision on what part of the map to move to pick-up orders within the $15 \times 15$ grid around its current location. The DDQN takes the state space tuple previously defined as its input. This includes the state of vehicles, supply, and demand. We feed the predicted requests in the next 15 minutes obtained from the demand model, the current location of each vehicle, and snapshots of vehicle location in the next 15, and 30 minutes. The network architecture of the DDQN consists of 16 convolution layers of $5 \times 5$, 32 convolution layer of $3 \times 3$, 64 convolution layer of $3 \times 3$, 16 convolution layer of $1 \times 1$. All convolution layers have a ReLu activation. The output of the deep Q network is an array of $15 \times 15$ Q values. Each value corresponds to the discounted sum of rewards a vehicle could get if dispatched to that particular zone.  As previously explained, Algorithm \ref{alg:Algo} shows detailed steps on how the FlexPool works.

In training this Q-network, we face the challenge of learning instability that is generally associated with Reinforcement Learning which uses rich function approximators such as Neural Networks. To combat this, we use experience replay and fixed Q-targets. This requires us to keep two sets of networks: the target network and the Q-network. The target network is what is ultimately used to generate the Q-values. The weights for the target network is periodically updated using the Q-network at an interval of 150 iterations. The Q-network on the other hand is learned at each iteration using a replay buffer of 10,000 steps. 

Given the distributed nature of our algorithm, every vehicle runs its own DDQN policy. As a result, the environment during training changes over time from the perspective of individual vehicles. To mitigate such effects, a new parameter $\beta$ is imposed to give the probability of performing an action in each time step. In our training procedure, $\beta$ is increased linearly from 0.3 to 1.0 over the first $T_n$ steps where $n$ is the number of vehicles. This allows only 30\% of the vehicles takes an actions at the onset. Consequently, the number of vehicles moving in each time step does not fluctuate significantly as our DDQN approaches the optimal policy.

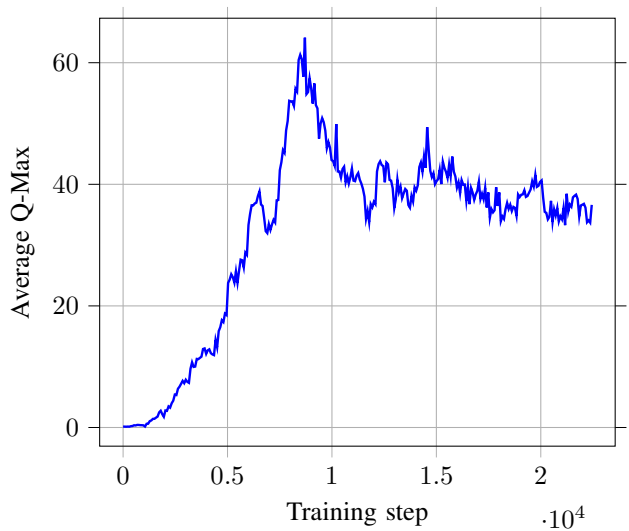
\begin{figure}
\begin{tikzpicture}

\begin{axis}[
tick align=outside,
x grid style={white!69.0196078431373!black},
xlabel={Training step},
xmajorgrids,
xmin=-1122, xmax=23562,
xtick style={color=white!69.0196078431373!black},
y grid style={white!69.0196078431373!black},
ylabel={Average Q-Max},
ymajorgrids,
ymin=-3.07121731743086, ymax=67.3191211140041,
ytick style={black}
]
\addplot [line width=0.9pt, blue]
table {%
	0 0.149212122
	75.0501672240803 0.141698318863934
	150.100334448161 0.128343520361638
	225.150501672241 0.149890086097368
	300.200668896321 0.134391917861099
	375.250836120401 0.214585461543176
	450.301003344482 0.246142712461059
	525.351170568562 0.364775630209777
	600.401337792642 0.338609105531161
	675.451505016722 0.423610686106841
	750.501672240803 0.417411903059881
	825.551839464883 0.383865966019961
	900.602006688963 0.376678985722963
	975.652173913043 0.351314161603199
	1050.70234113712 0.157906056324347
	1125.7525083612 0.607206934899323
	1200.80267558528 0.623817616046654
	1275.85284280936 1.0435193461814
	1350.90301003344 1.15811315094528
	1425.95317725753 1.41897896470618
	1501.00334448161 1.43040616768848
	1576.05351170569 1.6203934657751
	1651.10367892977 1.7959971657968
	1726.15384615385 2.48623692139848
	1801.20401337793 2.75658734694735
	1876.25418060201 2.18768480592351
	1951.30434782609 1.76303005922851
	2026.35451505017 2.79964854148971
	2101.40468227425 2.74463791918774
	2176.45484949833 3.46700724909454
	2251.50501672241 3.2665401998325
	2326.55518394649 3.98373530551264
	2401.60535117057 4.41545649935192
	2476.65551839465 5.40862711078577
	2551.70568561873 5.33734406826906
	2626.75585284281 6.37018301963726
	2701.80602006689 6.71645357591857
	2776.85618729097 7.19090563819404
	2851.90635451505 7.67011368555634
	2926.95652173913 7.2863766721537
	3002.00668896321 7.85857933995343
	3077.05685618729 7.50342327377622
	3152.10702341137 7.34917463082379
	3227.15719063545 9.6658077820443
	3302.20735785953 10.7180788957934
	3377.25752508361 9.96558180716002
	3452.30769230769 10.0218656313818
	3527.35785953177 11.2796007468355
	3602.40802675585 11.242295597105
	3677.45819397993 11.4499346446278
	3752.50836120401 11.6521740120967
	3827.55852842809 12.9115152143432
	3902.60869565217 13.0037847644154
	3977.65886287625 12.1111467180795
	4052.70903010033 12.6966123528094
	4127.75919732441 12.8549308645733
	4202.8093645485 12.2213284831425
	4277.85953177258 12.0164625873778
	4352.90969899666 11.9053384992569
	4427.95986622074 14.3988951760724
	4503.01003344482 13.1968123093381
	4578.0602006689 15.8527322530904
	4653.11036789298 16.497979117702
	4728.16053511706 17.6460585304627
	4803.21070234114 17.3450519774215
	4878.26086956522 18.7249875452712
	4953.3110367893 18.4952491527875
	5028.36120401338 23.7794007793329
	5103.41137123746 24.3279014053644
	5178.46153846154 25.2020371046231
	5253.51170568562 24.7267080282798
	5328.5618729097 23.8355748096111
	5403.61204013378 25.6749340896354
	5478.66220735786 23.8069446952259
	5553.71237458194 25.801204525443
	5628.76254180602 27.6072024868764
	5703.8127090301 27.5098745516664
	5778.86287625418 26.3210981515601
	5853.91304347826 28.7683317058144
	5928.96321070234 28.4018256297826
	6004.01337792642 33.3203979363805
	6079.0635451505 35.1040249381541
	6154.11371237458 36.541459887228
	6229.16387959866 36.5286720038715
	6304.21404682274 36.8461803697236
	6379.26421404682 36.9770619633454
	6454.3143812709 38.0629939415163
	6529.36454849498 38.8137419574053
	6604.41471571906 36.6747181794042
	6679.46488294314 36.4327508117021
	6754.51505016722 34.4758902268162
	6829.5652173913 32.291709720376
	6904.61538461538 31.9771714603283
	6979.66555183947 33.4470935318008
	7054.71571906355 32.50414158651
	7129.76588628763 33.4071623130464
	7204.81605351171 34.1607103959842
	7279.86622073579 33.7343302600284
	7354.91638795987 37.3657791805267
	7429.96655518395 37.4202042104513
	7505.01672240803 42.3304386448628
	7580.06688963211 43.6424192506869
	7655.11705685619 45.6346521163054
	7730.16722408027 45.1889949194388
	7805.21739130435 48.8560041315361
	7880.26755852843 50.4044625952237
	7955.31772575251 53.7381498574523
	8030.36789297659 53.6425540397743
	8105.41806020067 53.6485371739496
	8180.46822742475 52.898155026896
	8255.51839464883 55.7251461193958
	8330.56856187291 55.2744847995263
	8405.61872909699 60.4409848251469
	8480.66889632107 61.2872181093616
	8555.71906354515 60.5523448236225
	8630.76923076923 57.7051105225269
	8705.81939799331 64.1195602762116
	8780.86956521739 54.8350382332782
	8855.91973244147 55.1674773655471
	8930.96989966555 57.4281358239708
	9006.02006688963 55.7585354007093
	9081.07023411371 53.2878309432758
	9156.12040133779 56.5908789216541
	9231.17056856187 53.0272946863729
	9306.22073578595 52.4886736996959
	9381.27090301003 47.5144066488326
	9456.32107023411 49.9891920081499
	9531.37123745819 50.88506338711
	9606.42140468227 50.2152227010376
	9681.47157190636 48.8509053500967
	9756.52173913044 45.9267840490577
	9831.57190635451 46.9273943986682
	9906.6220735786 46.0431212550176
	9981.67224080268 43.9979165611139
	10056.7224080268 43.7887773708991
	10131.7725752508 43.0456687891673
	10206.8227424749 49.9142704130273
	10281.872909699 42.1229872539544
	10356.9230769231 42.1171581178145
	10431.9732441472 41.0971920816883
	10507.0234113712 42.5075128270209
	10582.0735785953 42.8876944401121
	10657.1237458194 40.3447756161141
	10732.1739130435 41.3125825176378
	10807.2240802676 40.1333941005433
	10882.2742474916 41.1596177965719
	10957.3244147157 42.5847685531654
	11032.3745819398 40.602169083351
	11107.4247491639 40.5174746300149
	11182.474916388 41.4423701167747
	11257.525083612 41.8617128854803
	11332.5752508361 40.7638441029994
	11407.6254180602 40.1283502672102
	11482.6755852843 39.266565650408
	11557.7257525084 37.9875105112788
	11632.7759197324 34.327030172387
	11707.8260869565 35.5993191722247
	11782.8762541806 33.7835541722498
	11857.9264214047 36.7452099606852
	11932.9765886288 36.2272239752554
	12008.0267558528 37.1389214835163
	12083.0769230769 36.7448701235983
	12158.127090301 42.1186221163031
	12233.1772575251 43.3156341018945
	12308.2274247492 43.7856541044884
	12383.2775919732 43.164238059768
	12458.3277591973 43.0067801543099
	12533.3779264214 40.3352696013011
	12608.4280936455 43.5822461838307
	12683.4782608696 43.2983011337835
	12758.5284280936 40.7198625093214
	12833.5785953177 40.6053086132319
	12908.6287625418 39.3564100028755
	12983.6789297659 36.0878328185486
	13058.72909699 37.5630657014989
	13133.779264214 40.7613404556911
	13208.8294314381 38.7417758012735
	13283.8795986622 39.5947547800938
	13358.9297658863 37.7280812631531
	13433.9799331104 38.5049749978309
	13509.0301003344 39.2341832084923
	13584.0802675585 38.9514712448992
	13659.1304347826 36.5977652467724
	13734.1806020067 36.1570384871321
	13809.2307692308 38.1051579826948
	13884.2809364549 37.1753278273888
	13959.3311036789 39.4918731321141
	14034.381270903 38.9623930927365
	14109.4314381271 39.8191083774372
	14184.4816053512 42.8478004667429
	14259.5317725753 43.5823784188228
	14334.5819397993 42.0933161827786
	14409.6321070234 44.5940822569858
	14484.6822742475 42.7201233050777
	14559.7324414716 49.4063434536695
	14634.7826086957 45.5402934578565
	14709.8327759197 42.197917754737
	14784.8829431438 41.184896967058
	14859.9331103679 41.8062088115761
	14934.983277592 40.0220786735215
	15010.0334448161 40.5158303429083
	15085.0836120401 40.8574073851232
	15160.1337792642 42.382944381875
	15235.1839464883 40.034786288327
	15310.2341137124 43.7247618490647
	15385.2842809365 42.7029505358157
	15460.3344481605 42.0510661881899
	15535.3846153846 41.0631313036341
	15610.4347826087 43.2229963919063
	15685.4849498328 41.3441928375388
	15760.5351170569 44.5634512396049
	15835.5852842809 42.0829021575274
	15910.635451505 41.2894523537175
	15985.6856187291 39.4623036023519
	16060.7357859532 40.3244103583572
	16135.7859531773 38.5594510056957
	16210.8361204013 40.6785417472982
	16285.8862876254 39.9422792351479
	16360.9364548495 39.9567561034774
	16435.9866220736 37.2712268752813
	16511.0367892977 39.5642079467094
	16586.0869565217 36.7322526102318
	16661.1371237458 39.3918258515087
	16736.1872909699 38.6470011890952
	16811.237458194 37.4493374468404
	16886.2876254181 38.3445026840053
	16961.3377926421 38.920500571045
	17036.3879598662 40.4304174818742
	17111.4381270903 37.5811725904545
	17186.4882943144 38.4479722826439
	17261.5384615385 37.5518276894247
	17336.5886287625 39.4073456764201
	17411.6387959866 36.1918678908666
	17486.6889632107 38.6568584084307
	17561.7391304348 34.518495416321
	17636.7892976589 35.9234613010464
	17711.8394648829 35.385689940239
	17786.889632107 35.7332157140731
	17861.9397993311 39.5263218356735
	17936.9899665552 36.436610451695
	18012.0401337793 38.6654708124681
	18087.0903010033 33.9346845561553
	18162.1404682274 34.7165213454323
	18237.1906354515 34.3051334641605
	18312.2408026756 35.916033072806
	18387.2909698997 36.9363206939935
	18462.3411371237 35.957009609935
	18537.3913043478 36.633608330507
	18612.4414715719 35.6072702335313
	18687.491638796 36.2880650540336
	18762.5418060201 36.130219085256
	18837.5919732441 34.8407784123414
	18912.6421404682 38.329308089736
	18987.6923076923 37.901105799375
	19062.7424749164 38.2755123645895
	19137.7926421405 38.3399501072137
	19212.8428093645 39.0051887231308
	19287.8929765886 37.9305831125229
	19362.9431438127 38.0783635598115
	19437.9933110368 38.6241328211809
	19513.0434782609 39.1522699372316
	19588.093645485 40.4948339752588
	19663.143812709 39.6389766293742
	19738.1939799331 41.5365721918251
	19813.2441471572 39.5793345632038
	19888.2943143813 39.7476216524045
	19963.3444816054 40.3828144578829
	20038.3946488294 40.6361992898706
	20113.4448160535 37.7975011637449
	20188.4949832776 35.4719829710699
	20263.5451505017 35.3032007612119
	20338.5953177258 34.2332012034614
	20413.6454849498 34.6415869194715
	20488.6956521739 37.2829361316719
	20563.745819398 33.8429777878978
	20638.7959866221 35.8698448627566
	20713.8461538462 34.8735114313289
	20788.8963210702 36.0999561311641
	20863.9464882943 33.8996144829164
	20938.9966555184 34.7458048167857
	21014.0468227425 34.2472133190981
	21089.0969899666 36.0689417727079
	21164.1471571906 33.2785128157071
	21239.1973244147 38.4475419810626
	21314.2474916388 35.5243267877031
	21389.2976588629 36.6818494428457
	21464.347826087 36.3223524138836
	21539.397993311 37.8513842514752
	21614.4481605351 38.1076027512448
	21689.4983277592 38.3052732155286
	21764.5484949833 37.5922260846818
	21839.5986622074 34.7173163645145
	21914.6488294314 36.5117987420317
	21989.6989966555 36.6454431322173
	22064.7491638796 36.7882687189729
	22139.7993311037 36.1571596873687
	22214.8494983278 33.7607176320945
	22289.8996655518 34.0706187996483
	22364.9498327759 33.6968938548012
	22440 36.61724842
};
\end{axis}

\end{tikzpicture}
\caption{This plot shows the convergence of Q-values during training. We see convergence in approximately 15,000 training steps.}\label{fig:Learning Curve}
\end{figure}

Training is performed using a total of 22,500 iterations, consisting of 30 episodes of 750 iterations each. Each iteration consists of 750 minutes of data, which is equivalent to two weeks data. The average q-max curve of all agents combined is plotted in figure \ref{fig:Learning Curve}. With the particular experiment shown in Figure \ref{fig:Learning Curve}, we initialized a fleet of $n=8000$ vehicles. The DDQN parameters included a start exploration rate $\epsilon = 1$ which was linearly decayed for $T_n$ steps which in this instance is 8000 steps.

It can be observed from the learning curve that over the first $T_{n}$ steps there is a gradual increase in the average Q-max of the fleet. This is explainable as it is within the exploration phase where the agents are gradually zeroing in on optimal policy. The average Q-max values hit a peak at 8000 steps and drop slightly as $\beta \to 1$. At this phase, all the agents are taking actions and competing for pick-ups, hence the average Q-max decreases until convergence at around 15,000 steps. In the experiment shown, the average Q-max converges at a value of approximately 35. 

We trained the DDQN agents in our simulation for a total of one month using all dataset requests from May. Consequently, we evaluated these trained DDQN agents by simulating two additional weeks from the month of June.

%% file: evaluations.tex
\section{Evaluation and Results}

In this section, we will evaluate the proposed approach, labeled as FlexPool w/ Hoptrips. We choose the parameters as $C_p=4$, $C_k=5$, $\beta_1=10$, $\beta_2=1$, $\beta_3=0.5$, $\beta_4=1$, and $\beta_5=1$, unless explicitly mentioned. We compare the proposed algorithm with two baselines as described below.
\begin{enumerate}
	\item \textbf{FlexPool w/o Hoptrips}: This baseline involves combined workloads for agents without hoptrips. %
	In this case, goods packages are delivered directly from pick-up to drop-off locations without transit. Since there are no hop-trips, $\beta_5=0$. 
	\item \textbf{Separate Vehicles}: This baseline is analogous to DeepPool \cite{al2019deeppool}, where passengers \& goods are pooled into separate sets of vehicles. Vehicle for passenger pick-ups is designated as ride-sharing vehicles with max capacity $C_p = 4$ assuming 4 seats in a car. Vehicles assigned to packages are designated as goods fulfillment vehicles with a max capacity $C_k = 10$ since we assume the entire car capacity (seats+trunk) can be used for goods. All other parameters are as in FlexPool w/o Hoptrips. %
\end{enumerate}

\begin{figure*}[ht]
	\begin{minipage}[t]{0.32\textwidth}
		\resizebox{\textwidth}{!}{  
			\input{fig_acceptrate}}
		\caption{This figure plots Accept Rates of all pick-up requests (passengers \& goods) for each of the 14 test days for all three models. We see that all three models accept above 90\% of all requests.}\label{fig: Overall Accept Rates}
	\end{minipage}
	\hspace{.01in}
	\begin{minipage}[t]{0.32\textwidth}
		\resizebox{\textwidth}{!}{  
			\input{fig_pass_accept} }
		\caption{This figure plots Accept Rates of passenger rideshare requests for each of the 14 test days for all three models. We see that all three models accept above 90\% of passenger rideshare requests.}\label{fig: Passenger Accept Rates}
	\end{minipage}
	\hspace{.01in}
	\begin{minipage}[t]{0.32\textwidth}
		\resizebox{\textwidth}{!}{  
			\input{fig_goods_accept}}
		\caption{This figure plots Accept Rates of goods delivery requests for each of the 14 test days for all three models. We see that all three models accept above 90\% of goods delivery requests.}\label{fig: Goods Accept Rates}
	\end{minipage}
\end{figure*}

\subsection{Evaluated Metrics}
For each of the considered algorithms, we evaluate the following metrics and outline their significance to the joint passenger and goods problem:
\begin{itemize}[leftmargin=*]
	\item \textbf{Accept Rate}: Accept rate is defined as the ratio of successful pick-ups by the fleet to the total number of requests made to the fleet in a given time slot. A high accept rate is a characteristic of a reliable mode of transportation. With a high accept rate, our fleet is able to fulfill the transportation demands for passenger and/or goods.
	\item \textbf{Fuel Cost per Delivery}: This is defined as the ratio of total fuel consumption by the fleet to the number of requests fulfilled by the fleet. The following assumptions are made in estimating cost:
	\begin{enumerate*}  
		\item\$2 (USD) per US gallon which is a ballpark estimate of US fuel prices as per \cite{gasprices}, and
		\item Each vehicle in the fleet consumes 0.5 US gallons per hour of driving.
	\end{enumerate*}
	 With a smaller amount of time a fleet vehicle spends traveling per request, less fuel is consumed which reduces congestion and emissions. A good performance on this metric, therefore, points towards a better overall transportation system efficiency. 
		{\item \textbf{Active Vehicles Ratio}:  The vehicles deployed from the fleet to serve customer demand are defined as ``active vehicles". We take an average of active vehicles in the fleet over the model evaluation duration. As defined in our reward in equation (\ref{eq:reward-5}), the objective of the fleet is to minimize the number of vehicles deployed on the road. By minimizing the number of active vehicles, we achieve better utilization of individual vehicles in serving the demand. 
			Given that (i) all baselines are catering to a similar volume of pickup orders, and (ii) all baselines are achieving a similar accept rate, a lower Active Vehicles Ratio indicates that a fleet is able to minimize the number of vehicles on the street to serve the requests. }
	\item \textbf{Wait time}: This is the time taken for a passenger or goods delivery request to be picked up by the fleet.  We note that wait time is an important metric for customer convenience with mobility-on-demand services. A low wait time as a result is ideal for both passengers and goods.
	\item \textbf{Effective Distance Ratio}: This is defined as the ratio of total distance covered if no multi-hops and sharing are allowed to the total distance covered when multi-hop and sharing is allowed. The efficient packing of vehicles alleviates the overall distance traveled by the vehicles in completing service for the same number of requests. Ride-sharing for passengers and multi-hop transport of goods reduces the effective distance of the vehicles due to efficient packing.

\end{itemize}

\begin{figure*}
	\begin{minipage}[t]{0.32\textwidth}
		\resizebox{\textwidth}{!}{  
			\input{fig_efficiency}}
		\caption{This figure plots Fuel Cost per Delivery with varying fleet size for all three models. We see that for all fleet sizes, FlexPool w/ Hoptrips has the lowest fuel cost in fulfilling requests.}\label{fig:Efficiency}
	\end{minipage}
	\hspace{.01in}
	\begin{minipage}[t]{0.32\textwidth}
		\resizebox{\textwidth}{!}{  
			\input{fig_active_vehicles}}
		\caption{This figure plots Active Vehicles Ratio with varying fleet size for all three models. We see that for all fleet sizes, FlexPool w/ Hoptrips achieves the lowest Active Vehicles Ratio in comparison to baselines.}\label{fig: Active vehicles}
	\end{minipage}
	\hspace{.01in}
	\begin{minipage}[t]{0.32\textwidth}
		\resizebox{\textwidth}{!}{  
			\input{fig_waittime}}
		\caption{This figure plots wait times of all requests for each of the 14 test days for all three models. We see that FlexPool w/ Hoptrips achieves lowest average wait time per pick-up in comparison to the baselines.}\label{fig:Wait time}
	\end{minipage}
\end{figure*}
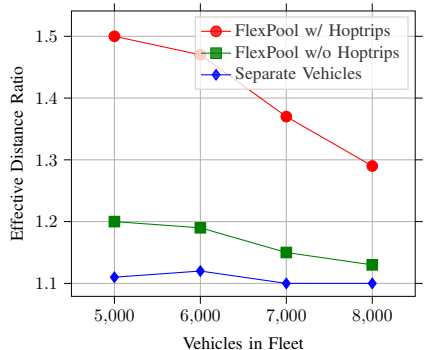
\begin{figure}
	\centering
	\resizebox{.32\textwidth}{!}{  
		\input{fig_eff_dist_ratio.tex}
	}
	\caption{This figure plots the effective distance ratio for all  three models. We see that the effective distance ratio of FlexPool is significantly larger than the baselines. With fewer fleet vehicles, FlexPool is able to achieve a better effective distance ratio thus achieving better packing of passengers and goods.}
	\label{eff_dist}
\end{figure}

\subsection{ Discussion on the Results}

In this section, we compare the results obtained from our simulation of our proposed algorithm (FlexPool w/ Hoptrips) against the two baseline scenarios defined previously (FlexPool w/o Hoptrips and Separate Vehicles).

In evaluating accept rates for the three models, we use a fleet size of $n=8000$ vehicles. Figures \ref{fig: Overall Accept Rates}, \ref{fig: Passenger Accept Rates}, and \ref{fig: Goods Accept Rates} show that the three algorithms accept  90\% of requests in the worst case with averages of 95\%.  In accepting passenger ride-share requests, FlexPool w/ hoptrips and FlexPool w/o hoptrips are identical in carrying capacities, however FlexPool w/ hoptrips is able to pick-up passengers at higher rates. In comparing goods delivery order accept rates (see Figure \ref{fig: Goods Accept Rates}), the Separate Vehicles baseline slightly outperforms other models potentially due to greater carrying capacity (both seats and trunk available for carrying goods). From the accept rate results, we observe that all three algorithms are in a similar ballpark. Therefore, we may compare the subsequent metrics to have a fair evaluation of system efficiency and sustainability across the three algorithms. 

We evaluate fuel cost per delivery across different fleet sizes for all three algorithms. Figure \ref{fig:Efficiency} shows the model performance on this metric. We observe from this plot that the proposed model (FlexPool w/ Hoptrips) clearly outperforms the baselines. With all three algorithms being in the same accept rate ballpark, this plot shows that the proposed model is able to fulfill delivery requests on an average with much lower fuel consumption. This points to cost savings on the part of the fleet operator. Additionally, given our simulator models fuel cost directly using vehicle travel time, a good performance on this metric indicates that vehicles spend less time on the road in fulfilling deliveries. This results in reducing congestion and emissions as well. With an average improvement of ~30\% from the next best baseline (FlexPool w/o Hoptrips), this goes to show the promise that using a combined workload with hoptrips has from both cost and sustainability standpoints. With both FlexPool models outperforming the separated vehicles model, it is evident that combining passenger \& goods workloads into one system is more sustainable and cost-efficient. 

In Fig. \ref{eff_dist}, we observe that the effective distance ratio varies with the number of fleet vehicles. Amongst all three algorithms, it is evident that our proposed algorithm achieves the highest effective distance ratio. This suggests that the vehicles are able to fulfill more deliveries at a given time as a result of being more efficiently packed.  The FlexPool w/o Hoptrips baseline performs second best while the Separate Vehicles baseline achieves the lowest effective distance ratio. This can be explained by the ability of FlexPool w/o Hoptrips to better utilize the full capacity of a vehicle (seating space + trunk space), while  Separate Vehicles underutilizes trunk space as a result of not considering goods requests while delivering passengers. 

In the definition of our global objective (refer to Section III), we included a component to minimize the number of deployed vehicles to serve passenger demand. Having noted that all 3 models achieve similar performance in accept rates for pickups, it is noteworthy that \ref{fig: Active vehicles} shows a significant improvement in the number of vehicles deployed with the FlexPool w/ Hoptrips. On average FlexPool w/ Hoptrips outperforms the next best baseline by approximately 35\%. This points to the higher utilization of fleet vehicles to achieve reliable accept rates. 

Figure \ref{fig:Wait time} shows the performance on the wait time of order pick-ups by the models over the test duration. It can be observed that FlexPool w/ hoptrips achieves the lowest time to pick-up orders. As there is a significant improvement in comparison to FlexPool w/o hoptrips, it is fair to infer that the use of a multi-hop routing for packages reduces the waiting time for orders to be picked up. Likewise, both FlexPool models outperform the Separate Vehicles scenario. We note that, when vehicles have the ability to pickup both passengers \& goods, requests are more likely to be accepted quickly resulting in improved customer wait times.

%% file: fig_acceptrate.tex
	\begin{tikzpicture}
	\begin{axis}[
	legend cell align={left},
	legend style={font=\small, fill opacity=0.7, draw opacity=1, text opacity=1, at={(0.03,0.03)}, anchor=south west, draw=white!80!black},
	tick align=outside,
	x grid style={white!69.0196078431373!black},
	xlabel={Date},
	xmajorgrids,
	xmin=-0.65, xmax=13.65,
	xtick style={color=black},
	xtick={0,1,2,3,4,5,6,7,8,9,10,11,12,13},
	xtick={0,1,2,3,4,5,6,7,8,9,10,11,12,13},
	xtick={0,1,2,3,4,5,6,7,8,9,10,11,12,13},
	xtick={0,1,2,3,4,5,6,7,8,9,10,11,12,13},
	xtick={0,1,2,3,4,5,6,7,8,9,10,11,12,13},
	xtick={0,1,2,3,4,5,6,7,8,9,10,11,12,13},
	xticklabels={06,07,08,09,10,11,12,13,14,15,16,17,18,19},
	xticklabels={06,07,08,09,10,11,12,13,14,15,16,17,18,19},
	xticklabels={06,07,08,09,10,11,12,13,14,15,16,17,18,19},
	xticklabels={06,07,08,09,10,11,12,13,14,15,16,17,18,19},
	xticklabels={06,07,08,09,10,11,12,13,14,15,16,17,18,19},
	xticklabels={06,07,08,09,10,11,12,13,14,15,16,17,18,19},
	y grid style={white!69.0196078431373!black},
	ylabel={Overall Accept Rate},	
	ymajorgrids,
	ymin=0.85, ymax=1,
	ytick style={color=black}
	]
	\addplot [line width=0.7pt, red, opacity=0.5, mark=*, mark size=2.5, mark options={solid}]
	table {%
		0 0.944619211526163
		1 0.964564213262772
		2 0.954840938422029
		3 0.991368108409133
		4 0.990574707730449
		5 0.992966717381323
		6 0.951791175224712
		7 0.953602672104902
		8 0.932177068804899
		9 0.958608561379429
		10 0.958441175851693
		11 0.935854382704041
		12 0.975246841335976
		13 0.950492354592084
	};
	\addlegendentry{FlexPool w/ Hoptrips}
	\addplot [line width=1.5pt, red, dashed, forget plot]
	table {%
		0 0.961082009194972
		1 0.961082009194972
		2 0.961082009194972
		3 0.961082009194972
		4 0.961082009194972
		5 0.961082009194972
		6 0.961082009194972
		7 0.961082009194972
		8 0.961082009194972
		9 0.961082009194972
		10 0.961082009194972
		11 0.961082009194972
		12 0.961082009194972
		13 0.961082009194972
	};
	\addplot [line width=0.7pt, green!50.1960784313725!black, opacity=0.5, mark=square*, mark size=2.5, mark options={solid}]
	table {%
		0 0.949033756655146
		1 0.997625188203835
		2 0.925438799610985
		3 0.995992194799372
		4 0.906711574631578
		5 0.898195930005818
		6 0.950366092275174
		7 0.976831847868646
		8 0.96
		9 0.901225924495754
		10 0.903468464605754
		11 0.979405857396373
		12 0.990698922740903
		13 0.973751540088209
	};
	\addlegendentry{FlexPool w/o Hoptrips}
	\addplot [line width=1.5pt, green!50.1960784313725!black, dashed, forget plot]
	table {%
		0 0.950624720955539
		1 0.950624720955539
		2 0.950624720955539
		3 0.950624720955539
		4 0.950624720955539
		5 0.950624720955539
		6 0.950624720955539
		7 0.950624720955539
		8 0.950624720955539
		9 0.950624720955539
		10 0.950624720955539
		11 0.950624720955539
		12 0.950624720955539
		13 0.950624720955539
	};
	\addplot [line width=0.7pt, blue, opacity=0.5, mark=diamond*, mark size=2.5, mark options={solid}]
	table {%
		0 0.977074022295785
		1 0.951105993612147
		2 0.935668909651768
		3 0.989807391217544
		4 0.945819437197896
		5 0.989584419682372
		6 0.982865783710502
		7 0.972345584536743
		8 0.960468499811153
		9 0.960989363307182
		10 0.990990539821323
		11 0.983263589479004
		12 0.964997769273366
		13 0.974708840342513
	};
	\addlegendentry{Separate Vehicles}
	\addplot [line width=1.5pt, blue, dashed, forget plot]
	table {%
		0 0.969977867424235
		1 0.969977867424235
		2 0.969977867424235
		3 0.969977867424235
		4 0.969977867424235
		5 0.969977867424235
		6 0.969977867424235
		7 0.969977867424235
		8 0.969977867424235
		9 0.969977867424235
		10 0.969977867424235
		11 0.969977867424235
		12 0.969977867424235
		13 0.969977867424235
	};
	\end{axis}
	
	\end{tikzpicture}

%% file: fig_pass_accept.tex
	\begin{tikzpicture}
	
	\begin{axis}[
	legend cell align={left},
	legend style={font=\small, fill opacity=0.7, draw opacity=1, text opacity=1, at={(0.03,0.03)}, anchor=south west, draw=white!80!black},
	tick align=outside,
	x grid style={white!69.0196078431373!black},
	xlabel={Date},
	xmajorgrids,
	xmin=-0.65, xmax=13.65,
	xtick style={color=black},
	xtick={0,1,2,3,4,5,6,7,8,9,10,11,12,13},
	xtick={0,1,2,3,4,5,6,7,8,9,10,11,12,13},
	xtick={0,1,2,3,4,5,6,7,8,9,10,11,12,13},
	xtick={0,1,2,3,4,5,6,7,8,9,10,11,12,13},
	xtick={0,1,2,3,4,5,6,7,8,9,10,11,12,13},
	xtick={0,1,2,3,4,5,6,7,8,9,10,11,12,13},
	xticklabels={06,07,08,09,10,11,12,13,14,15,16,17,18,19},
	xticklabels={06,07,08,09,10,11,12,13,14,15,16,17,18,19},
	xticklabels={06,07,08,09,10,11,12,13,14,15,16,17,18,19},
	xticklabels={06,07,08,09,10,11,12,13,14,15,16,17,18,19},
	xticklabels={06,07,08,09,10,11,12,13,14,15,16,17,18,19},
	xticklabels={06,07,08,09,10,11,12,13,14,15,16,17,18,19},
	y grid style={white!69.0196078431373!black},
	ylabel={Passenger Accept Rate},	
	ymajorgrids,
	ymin=0.85, ymax=1,
	ytick style={color=black}
	]
	\addplot [line width=0.7pt, red, opacity=0.5, mark=*, mark size=2.5, mark options={solid}]
	table {%
		0 0.992520131005442
		1 0.976253264646995
		2 0.990489016716504
		3 0.996920637728031
		4 0.994841508660313
		5 0.993611093655083
		6 0.998645614720094
		7 0.98954365705136
		8 0.98650777054534
		9 0.994063772048847
		10 0.997310904620725
		11 0.991311093198818
		12 0.985439659238986
		13 0.985592227151593
	};
	\addlegendentry{FlexPool w/ Hoptrips}
	\addplot [line width=1.5pt, red, dashed, forget plot]
	table {%
		0 0.990932167927723
		1 0.990932167927723
		2 0.990932167927723
		3 0.990932167927723
		4 0.990932167927723
		5 0.990932167927723
		6 0.990932167927723
		7 0.990932167927723
		8 0.990932167927723
		9 0.990932167927723
		10 0.990932167927723
		11 0.990932167927723
		12 0.990932167927723
		13 0.990932167927723
	};
	\addplot [line width=0.7pt, green!50.1960784313725!black, opacity=0.5, mark=square*, mark size=2.5, mark options={solid}]
	table {%
		0 0.96731195908067
		1 0.996693274782252
		2 0.906621571856097
		3 0.994103190364233
		4 0.896011359982078
		5 0.876426513833043
		6 0.94079863308442
		7 0.986622557260878
		8 0.96
		9 0.899694944185771
		10 0.909823058321849
		11 0.976683975601171
		12 0.984538818334961
		13 0.988201286059819
	};
	\addlegendentry{FlexPool w/o Hoptrips}
	\addplot [line width=1.5pt, green!50.1960784313725!black, dashed, forget plot]
	table {%
		0 0.948823653053375
		1 0.948823653053375
		2 0.948823653053375
		3 0.948823653053375
		4 0.948823653053375
		5 0.948823653053375
		6 0.948823653053375
		7 0.948823653053375
		8 0.948823653053375
		9 0.948823653053375
		10 0.948823653053375
		11 0.948823653053375
		12 0.948823653053375
		13 0.948823653053375
	};
	\addplot [line width=0.7pt, blue, opacity=0.5, mark=diamond*, mark size=2.5, mark options={solid}]
	table {%
		0 0.967243070497405
		1 0.988062105451064
		2 0.932145289357444
		3 0.99814737772693
		4 0.959188614002696
		5 0.976589504556405
		6 0.986912598551619
		7 0.961263873577194
		8 0.982358882180469
		9 0.984011877959205
		10 0.990009056830597
		11 0.979089327514466
		12 0.967385368029969
		13 0.954752810003742
	};
	\addlegendentry{Separate Vehicles}
	\addplot [line width=1.5pt, blue, dashed, forget plot]
	table {%
		0 0.973368554017086
		1 0.973368554017086
		2 0.973368554017086
		3 0.973368554017086
		4 0.973368554017086
		5 0.973368554017086
		6 0.973368554017086
		7 0.973368554017086
		8 0.973368554017086
		9 0.973368554017086
		10 0.973368554017086
		11 0.973368554017086
		12 0.973368554017086
		13 0.973368554017086
	};
	\end{axis}
	
	\end{tikzpicture}

%% file: fig_goods_accept.tex
	\begin{tikzpicture}
	
	\begin{axis}[
	legend cell align={left},
	legend style={font=\small, fill opacity=0.7, draw opacity=1, text opacity=1, at={(0.03,0.03)}, anchor=south west, draw=white!80!black},
	tick align=outside,
	x grid style={white!69.0196078431373!black},
	xlabel={Date},
	xmajorgrids,
	xmin=-0.65, xmax=13.65,
	xtick style={color=black},
	xtick={0,1,2,3,4,5,6,7,8,9,10,11,12,13},
	xtick={0,1,2,3,4,5,6,7,8,9,10,11,12,13},
	xtick={0,1,2,3,4,5,6,7,8,9,10,11,12,13},
	xtick={0,1,2,3,4,5,6,7,8,9,10,11,12,13},
	xtick={0,1,2,3,4,5,6,7,8,9,10,11,12,13},
	xtick={0,1,2,3,4,5,6,7,8,9,10,11,12,13},
	xticklabels={06,07,08,09,10,11,12,13,14,15,16,17,18,19},
	xticklabels={06,07,08,09,10,11,12,13,14,15,16,17,18,19},
	xticklabels={06,07,08,09,10,11,12,13,14,15,16,17,18,19},
	xticklabels={06,07,08,09,10,11,12,13,14,15,16,17,18,19},
	xticklabels={06,07,08,09,10,11,12,13,14,15,16,17,18,19},
	xticklabels={06,07,08,09,10,11,12,13,14,15,16,17,18,19},
	y grid style={white!69.0196078431373!black},
	ylabel={Goods Accept Rate},	
	ymajorgrids,
	ymin=0.85, ymax=1,
	ytick style={color=black}
	]
	\addplot [line width=0.7pt, red, opacity=0.5, mark=*, mark size=2.5, mark options={solid}]
	table {%
		0 0.915022138368608
		1 0.934781001068288
		2 0.90737457448533
		3 0.980916257446227
		4 0.982776958592183
		5 0.991797671678849
		6 0.924718629002011
		7 0.918932454633534
		8 0.896360671316529
		9 0.932337674164466
		10 0.929914524244666
		11 0.908828455152343
		12 0.937306868401129
		13 0.912273682525911
	};
	\addlegendentry{FlexPool w/ Hoptrips}
	\addplot [line width=1.5pt, red, dashed, forget plot]
	table {%
		0 0.933810111505719
		1 0.933810111505719
		2 0.933810111505719
		3 0.933810111505719
		4 0.933810111505719
		5 0.933810111505719
		6 0.933810111505719
		7 0.933810111505719
		8 0.933810111505719
		9 0.933810111505719
		10 0.933810111505719
		11 0.933810111505719
		12 0.933810111505719
		13 0.933810111505719
	};
	\addplot [line width=0.7pt, green!50.1960784313725!black, opacity=0.5, mark=square*, mark size=2.5, mark options={solid}]
	table {%
		0 0.932740503653602
		1 1
		2 0.941203380273699
		3 0.999633342377555
		4 0.915956941981568
		5 0.914360802575729
		6 0.961309750015382
		7 0.966533402421575
		8 0.96
		9 0.902367124806309
		10 0.899011452957161
		11 0.983621995350732
		12 1
		13 0.958167869318332
	};
	\addlegendentry{FlexPool w/o Hoptrips}
	\addplot [line width=1.5pt, green!50.1960784313725!black, dashed, forget plot]
	table {%
		0 0.952493326123689
		1 0.952493326123689
		2 0.952493326123689
		3 0.952493326123689
		4 0.952493326123689
		5 0.952493326123689
		6 0.952493326123689
		7 0.952493326123689
		8 0.952493326123689
		9 0.952493326123689
		10 0.952493326123689
		11 0.952493326123689
		12 0.952493326123689
		13 0.952493326123689
	};
	\addplot [line width=0.7pt, blue, opacity=0.5, mark=diamond*, mark size=2.5, mark options={solid}]
	table {%
		0 0.99022957844283
		1 0.915872776430453
		2 0.942233405393912
		3 0.981954945507812
		4 0.922447445550558
		5 0.997649011501426
		6 0.978920550969634
		7 0.989418639578384
		8 0.927929923879185
		9 0.937569905727134
		10 0.991909538948297
		11 0.98743017939986
		12 0.964015630932793
		13 0.986309488087379
	};
	\addlegendentry{Separate Vehicles}
	\addplot [line width=1.5pt, blue, dashed, forget plot]
	table {%
		0 0.965277930024976
		1 0.965277930024976
		2 0.965277930024976
		3 0.965277930024976
		4 0.965277930024976
		5 0.965277930024976
		6 0.965277930024976
		7 0.965277930024976
		8 0.965277930024976
		9 0.965277930024976
		10 0.965277930024976
		11 0.965277930024976
		12 0.965277930024976
		13 0.965277930024976
	};
	\end{axis}
	
	\end{tikzpicture}

%% file: fig_efficiency.tex
	\begin{tikzpicture}
	
	\begin{axis}[
	legend cell align={left},
	legend style={font=\small, fill opacity=0.7, draw opacity=1, text opacity=1, at={(0.97,0.03)}, anchor=south east, draw=white!80!black},
	tick align=outside,
	x grid style={white!69.0196078431373!black},
	xmajorgrids,
	xlabel={Fleet Size},
	xmin=4500, xmax=8500,
	xtick style={color=black},
	y grid style={white!69.0196078431373!black},
	ymajorgrids,
	ylabel={Fuel Cost per Delivery (\$)},
	ymin=0.0004399375, ymax=0.0011013125,
	ytick style={color=black}
	]
	\addplot [semithick, red, mark=*, mark size=3, mark options={solid}]
	table {%
		5000 0.00047
		6000 0.000545
		7000 0.00063
		8000 0.00068125
	};
	\addlegendentry{FlexPool w/ Hoptrips}
	\addplot [semithick, green!50.1960784313725!black, mark=square*, mark size=3, mark options={solid}]
	table {%
		5000 0.000654
		6000 0.000633333333333333
		7000 0.000948571428571429
		8000 0.001005
	};
	\addlegendentry{FlexPool w/o Hoptrips}
	\addplot [semithick, blue, mark=diamond*, mark size=3, mark options={solid}]
	table {%
		5000 0.000688
		6000 0.000935
		7000 0.000985714285714286
		8000 0.00107125
	};
	\addlegendentry{Separate Vehicles}
	\end{axis}
	
	\end{tikzpicture}

%% file: fig_active_vehicles.tex
	\begin{tikzpicture}
	
	\begin{axis}[
	legend cell align={left},
	legend style={font=\small, fill opacity=0.8, draw opacity=1, text opacity=1, at={(0.03,0.5)}, anchor=west, draw=white!80!black},
	tick align=outside,
	x grid style={white!69.0196078431373!black},
	xlabel={Vehicles in Fleet},
	xmajorgrids,
	xmin=4500, xmax=8500,
	xtick style={color=black},
	y grid style={white!69.0196078431373!black},
	ylabel={Active Vehicle Ratio},
	ymajorgrids,
	ymin=0.3554828125, ymax=0.6528609375,
	ytick style={color=black}
	]
	\addplot [semithick, red, mark=*, mark size=3, mark options={solid}]
	table {%
		5000 0.369
		6000 0.391
		7000 0.377428571428571
		8000 0.37925
	};
	\addlegendentry{FlexPool w/ Hoptrips}
	\addplot [semithick, green!50.1960784313725!black, mark=square*, mark size=3, mark options={solid}]
	table {%
		5000 0.6218
		6000 0.615458333333333
		7000 0.623625
		8000 0.63934375
	};
	\addlegendentry{FlexPool w/o Hoptrips}
	\addplot [semithick, blue, mark=diamond*,  mark size=3, mark options={solid}]
	table {%
		5000 0.6002
		6000 0.5725
		7000 0.493428571428571
		8000 0.622375
	};
	\addlegendentry{Separate Vehicles}
	\end{axis}
	
	\end{tikzpicture}

%% file: fig_waittime.tex
	\begin{tikzpicture}
	
	\begin{axis}[
	legend cell align={left},
	legend style={fill opacity=0.8, draw opacity=1, text opacity=1, draw=white!80!black},
	tick align=outside,
	x grid style={white!69.0196078431373!black},
	xlabel={Date},
	xmajorgrids,
	xmin=-0.65, xmax=13.65,
	xtick style={color=black},
	y grid style={white!69.0196078431373!black},
	ylabel={Wait Time (mins)},
	ymajorgrids,
	ymin=1.29998911684233, ymax=9.01060769252194,
	ytick style={color=black}
	]
	\addplot [semithick, red, opacity=0.7, mark=*, mark size=2.5, mark options={solid}]
	table {%
		0 4.30952681407953
		1 3.61563300741756
		2 4.09287328075063
		3 3.11377707283825
		4 4.68606126075353
		5 3.30270169892383
		6 3.30687391982883
		7 5.49649823869041
		8 3.62321022550485
		9 3.1513477540558
		10 3.00469352312228
		11 3.05620146202198
		12 4.41087040460081
		13 4.6588979457235
	};
	\addlegendentry{FlexPool w/ Hoptrips}
	\addplot [line width=1.5pt, red, dashed, forget plot]
	table {%
		0 3.84494047202227
		1 3.84494047202227
		2 3.84494047202227
		3 3.84494047202227
		4 3.84494047202227
		5 3.84494047202227
		6 3.84494047202227
		7 3.84494047202227
		8 3.84494047202227
		9 3.84494047202227
		10 3.84494047202227
		11 3.84494047202227
		12 3.84494047202227
		13 3.84494047202227
	};
	\addplot [semithick, green!50.1960784313725!black, opacity=0.7, mark=square*, mark size=2.5, mark options={solid}]
	table {%
		0 6.32551953242181
		1 2.91182249452067
		2 6.27759057022306
		3 3.72795307836005
		4 6.11295038939893
		5 5.81619813676731
		6 6.30791618918044
		7 6.06370162086475
		8 2.58222907203021
		9 6.56710240933699
		10 6.13108664834842
		11 4.66578324072386
		12 1.65047177937322
		13 6.12296818081204
	};
	\addlegendentry{FlexPool w/o Hoptrips}
	\addplot [line width=1.5pt, green!50.1960784313725!black, dashed, forget plot]
	table {%
		0 5.09023523874013
		1 5.09023523874013
		2 5.09023523874013
		3 5.09023523874013
		4 5.09023523874013
		5 5.09023523874013
		6 5.09023523874013
		7 5.09023523874013
		8 5.09023523874013
		9 5.09023523874013
		10 5.09023523874013
		11 5.09023523874013
		12 5.09023523874013
		13 5.09023523874013
	};
	\addplot [semithick, blue, opacity=0.5, mark=diamond*, mark size=2.5, mark options={solid}]
	table {%
		0 5.4358322903243
		1 7.23942646128044
		2 8.66012502999105
		3 5.80489554008094
		4 6.88077142293624
		5 5.04208765756356
		6 4.90025364701032
		7 5.4966403177753
		8 7.11664957090823
		9 6.9446085016215
		10 4.25953982552275
		11 4.48748117399941
		12 6.11288095240628
		13 5.5293555467371
	};
	\addlegendentry{Separate Vehicles}
	\addplot [line width=1.5pt, blue, dashed, forget plot]
	table {%
		0 5.99361056701124
		1 5.99361056701124
		2 5.99361056701124
		3 5.99361056701124
		4 5.99361056701124
		5 5.99361056701124
		6 5.99361056701124
		7 5.99361056701124
		8 5.99361056701124
		9 5.99361056701124
		10 5.99361056701124
		11 5.99361056701124
		12 5.99361056701124
		13 5.99361056701124
	};
	\end{axis}
	
	\end{tikzpicture}

%% file: fig_eff_dist_ratio.tex
\begin{tikzpicture}

\begin{axis}[
legend cell align={left},
legend style={fill opacity=0.8, draw opacity=1, text opacity=1, draw=white!80!black},
tick align=outside,
x grid style={white!69.0196078431373!black},
xlabel={Vehicles in Fleet},
xmajorgrids,
xmin=4500, xmax=8500,
xtick style={color=black},
y grid style={white!69.0196078431373!black},
ylabel={Effective Distance Ratio},
ymajorgrids,
ymin=1.079, ymax=1.541,
ytick style={color=black}
]
\addplot [semithick, red, mark=*, mark size=3, mark options={solid}]
table {%
5000 1.5
6000 1.47
7000 1.37
8000 1.29
};
\addlegendentry{FlexPool w/ Hoptrips}
\addplot [semithick, green!50.1960784313725!black, mark=square*, mark size=3, mark options={solid}]
table {%
5000 1.2
6000 1.19
7000 1.15
8000 1.13
};
\addlegendentry{FlexPool w/o Hoptrips}
\addplot [semithick, blue, mark=diamond*,  mark size=3, mark options={solid}]
table {%
5000 1.11
6000 1.12
7000 1.1
8000 1.1
};
\addlegendentry{Separate Vehicles}
\end{axis}

\end{tikzpicture}

%% file: conclusions.tex
\section{Conclusions and Future Work}

We propose FlexPool as a distributed model-free algorithm for joint ride-sharing of passengers and goods, which uses deep neural networks and reinforcement learning to learn optimal dispatch policies by interacting with the environment and an efficient matching of the passengers and goods to the vehicles. The proposed approach enables pooling of passengers as well as multi-hop transfer of goods, helping efficient use of the vehicles. Through efficiently incorporating passenger and goods delivery demand statistics and deep learning models, our proposed method manages dispatching and matching solutions for an efficient and sustainable combined transportation service. In addressing the problem of joint transportation of passengers and goods workloads, FlexPool is able to achieve a significantly higher operational efficiency as well as a lower environmental footprint. As a distributed system, FlexPool can adapt fluidly to a dynamic environment with fluctuations in demands of different workloads. 

FlexPool assumes existing infrastructure to enable convenient storage during multi-hop transit for goods delivery. In practice however, cost efficient methods for goods transit are currently not present and this would be very valuable piece of infrastructure to allow multi-hop transit. Along the same lines, research of practical incentives to encourage multi-hop transfers such as new pricing models would be of great use in enabling this technology. Another important aspect that needs to be explored is the incorporation of deadline based constraints to allow for delivery of urgent goods or service-based pricing contracts with customers. Likewise, a multi-agent formulation involving coordination among vehicles would be an interesting extension to the proposed solution to the shared passengers \& goods delivery problem.

%% file: ETA.tex
\begin{figure}[htbp]
	\includegraphics[width=7.7cm]{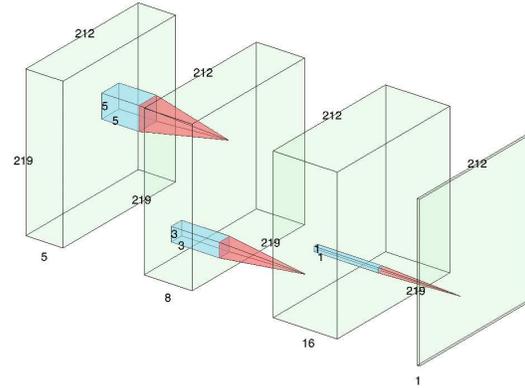}
	\caption{This diagram shows the Conv-Net architecture of the Demand Prediction Model.}
	\label{fig:Demand}
\end{figure}

\begin{figure*}[htbp]
	\begin{subfigure}{.48\textwidth}
		\centering
		\includegraphics[width=7.7cm]{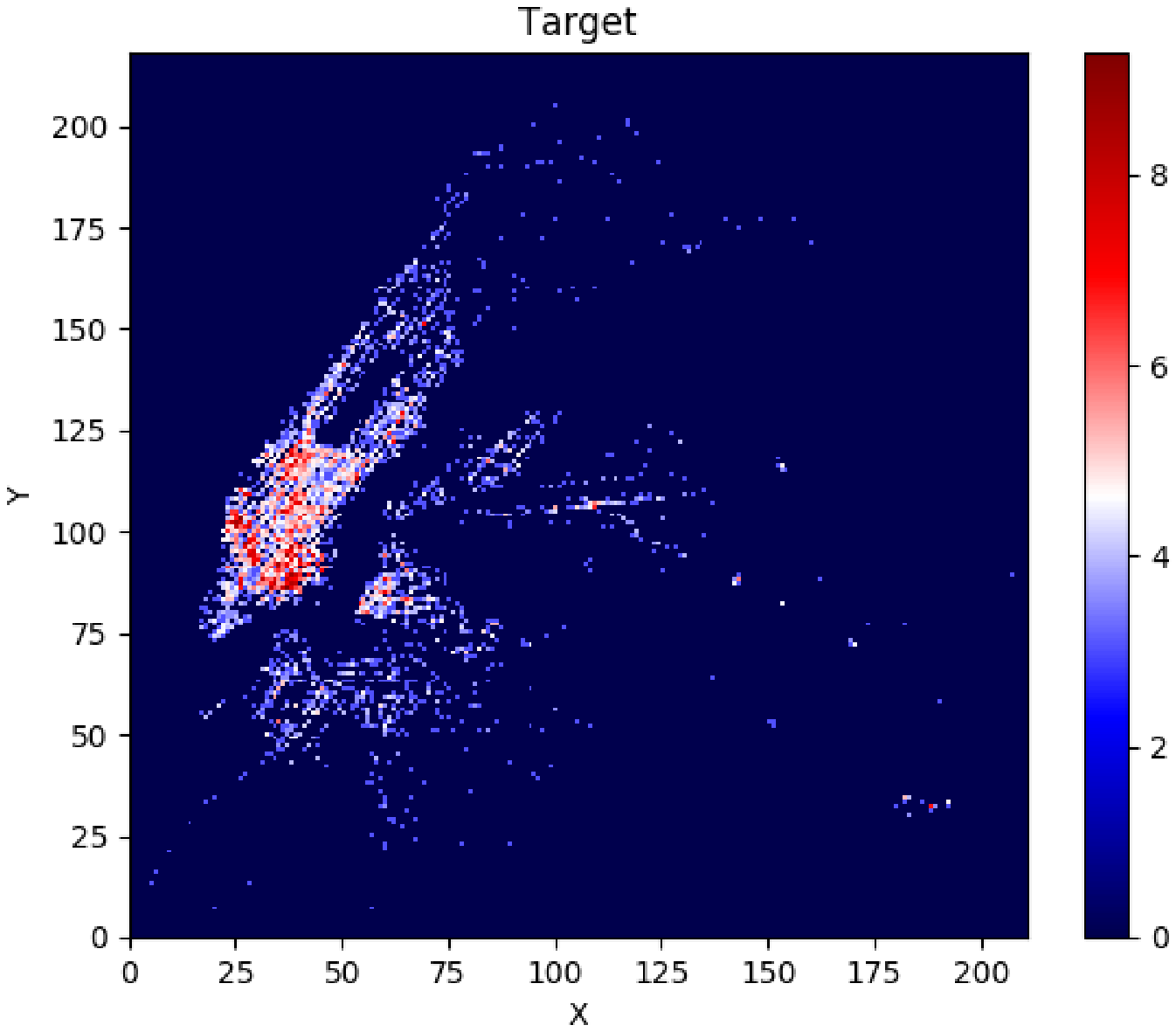}
		\caption{This shows a heatmap of actual demand from the dataset which is used as a target to train the demand prediction model.} \label{fig:1a}
	\end{subfigure}
	\begin{subfigure}{.48\textwidth}
		\centering
		\includegraphics[width=7.7cm]{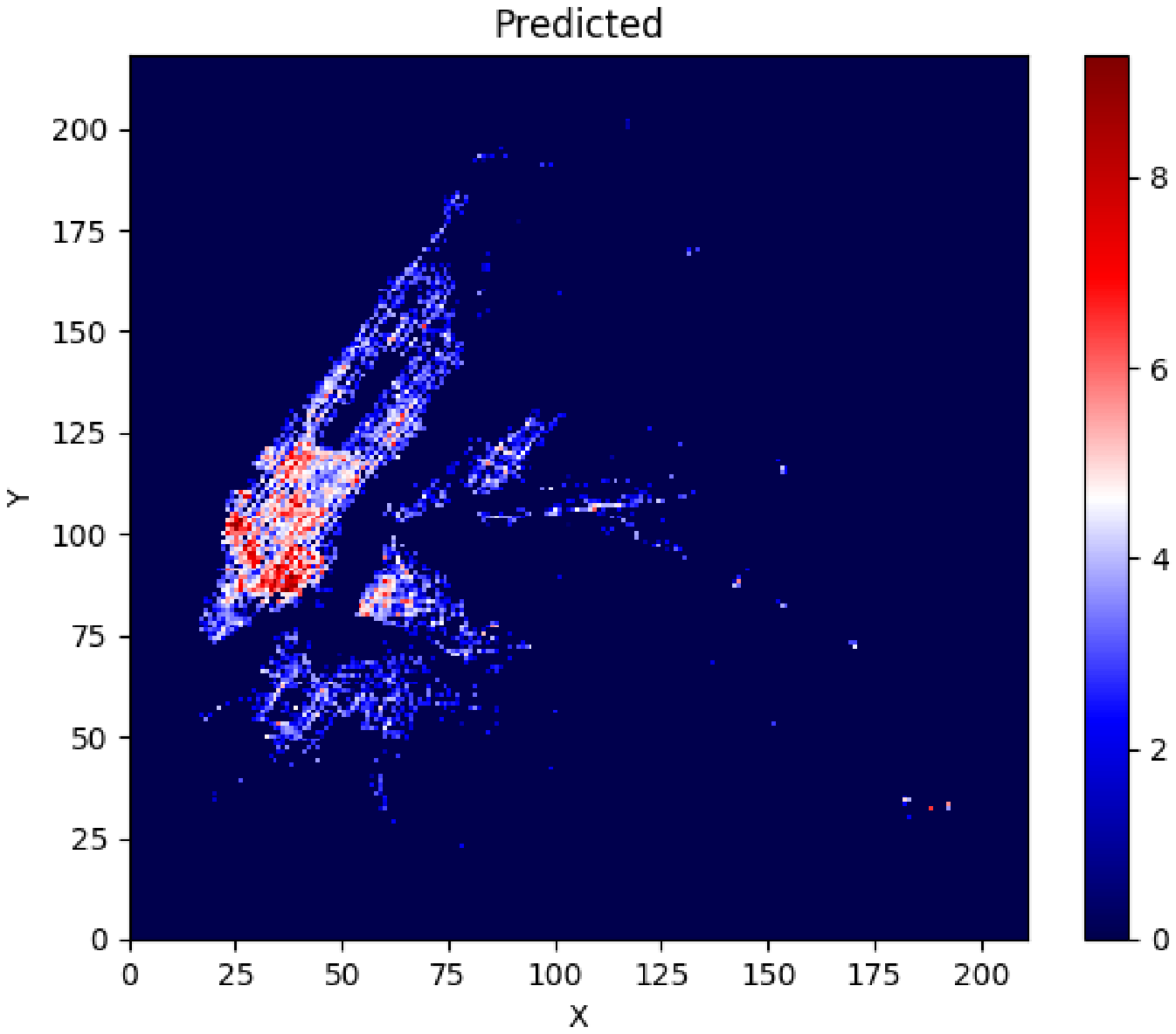}
		\caption{This shows a heatmap obtained from the demand prediction model which predicted pickup demand over the map.} \label{fig:1b}
	\end{subfigure}
	\caption{The above heatmaps visualize target and predicted demand areas. Warmer colors indicate a higher demand as shown by the color scale.} \label{fig:Demand_Heatmap}
\end{figure*}

\section{ETA and Demand Prediction}\label{eta_pred}

The simulator uses an Estimated Time of Arrival (ETA) model to predict the estimated trip times. This model is built using the New York City taxi data set. In the ETA model, we want to predict the expected travel time between two zones (two pairs of latitudes and longitudes). We split our data into 70\% train and 30\% test. We use day of week, latitude, longitude and time of days as the explanatory variables and use random forest to predict the ETA. The final ETA model yielded a root mean squared error (RMSE) of 3.4 on the test data.

The demand prediction model is a critical element to the simulator in building the state space vector that allows DDQN agents to proactively dispatch towards areas where there is a high demand. This model is built using the Conv-Net architecture shown in figure \ref{fig:Demand}. The network outputs a  $212 \times 219$  heat  map image in which each pixel stands for the predicted number of pick-up requests for each location on the map for the following 30 minutes of simulation. The network is fed with input images which represent the actual pick-ups of all service types over the last 6 time-steps. The actual pick-up counts over the map is combined with the sine and cosine of the day of week and hour of day to capture the daily and weekly periodicity of the demand. After training using a 80\% train and 20\% test split, RMSE values for training and testing were 0.945 and 1.217 respectively. Figure \ref{fig:Demand_Heatmap} shows the demand heat map of a target sample and a predicted sample of this demand prediction model. The predicted demand in this figure is a sample $212 \times 219$ output of the Conv-Net Architecture from Figure \ref{fig:Demand}.

%% file: notations.tex
\begin{color}{blue} 
	\begin{table*}[t]
\begin{tabular}{ |p{5cm}||p{10cm}|  }
\hline
\textbf{Symbol} & \textbf{Description} \\
\hline
$N$   & number of vehicles    \\
\hline
${v}_{t,i}$ &  number of available vehicles at time $t$ in map region $i$ \\
\hline
${d}_{t,i}$ & number of requests at time $t$ in map region $i$ \\
\hline
$d_{t,\widetilde{t},i}$ & number of additional vehicles anticipated to become available in region $i$ by time $\widetilde{t}$ at time $t$ \\
\hline
${x}_{t,n}$ & current location for vehicle $n$, current status, available seating, available trunk space, time at which a request is picked up, and destination of each request\\
\hline
$\varOmega_{t,n}$ &  state of vehicle $n$ at time $t$ \\
\hline
${a}_{t,n}$ & action of vehicle $n$ at time $t$ \\
\hline
$r_{t,n}$ & reward of vehicle $n$ at time $t$ \\
\hline
$\beta_i$ & weight of component $i$ in the reward expression \\
\hline
$b_{t,n}$ & number of customers served by vehicle $n$ at time $t$ \\
\hline
$p_{t,n}$ & number of packages being carried by vehicle $n$ at time $t$ \\
\hline
$c_{t,n}$ & time spent by vehicle $n$ to hop or take a detour to pick up additional requests \\
\hline
$\omega_u$ & delivery ``urgency" weight assigned for the $u$'th request \\ 
\hline
$\delta_{t,n,u}$ & additional driving time spent on delivering request $u$ by vehicle $n$ due to other concurrent deliveries \\
\hline
$e_{t,n}$ & occupied flag for vehicle $n$ at time $t$ \\ 
\hline
$H_u$ & number of ``hops" completed by package $u$ \\
\hline
$\mathcal{L}_{t,n}$ & list of all requests assigned to vehicle $n$ at time $t$ \\
\hline

\end{tabular}
\caption{Table of notations used in the paper. }\label{tbl_not}
\end{table*}
\end{color}

%% file: joint_pass_good.bbl
\begin{thebibliography}{10}

\bibitem{asdecker2020drives}
B.~Asdecker and F.~Zirkelbach, ``What drives the drivers? a qualitative
  perspective on what motivates the crowd delivery workforce,'' in {\em
  Proceedings of the 53rd Hawaii International Conference on System Sciences},
  2020.

\bibitem{arora2019m}
S.~Arora and A.~Verma, ``M-commerce: Crusader for “phygital” retail,'' {\em
  M-Commerce: Experiencing the Phygital Retail}, p.~163, 2019.

\bibitem{jo2019impact}
Y.~J. Jo, M.~Matsumura, and D.~E. Weinstein, ``The impact of e-commerce on
  relative prices and consumer welfare,'' tech. rep., National Bureau of
  Economic Research, 2019.

\bibitem{bubner2014logistics}
N.~Bubner, N.~Bubner, R.~Helfifg, and M.~Jeske, ``Logistics trend radar,'' {\em
  DHL Trend Research}, 2014.

\bibitem{yaraghi2017current}
N.~Yaraghi and S.~Ravi, ``The current and future state of the sharing
  economy,'' {\em Available at SSRN 3041207}, 2017.

\bibitem{hahn2017ridesharing}
R.~Hahn and R.~Metcalfe, ``The ridesharing revolution: Economic survey and
  synthesis,'' {\em More equal by design: economic design responses to
  inequality}, vol.~4, 2017.

\bibitem{kokalitcheva2016uber}
K.~Kokalitcheva, ``Uber now has 40 million monthly riders worldwide,'' {\em
  Fortune Magazine}, 2016.

\bibitem{crowddeliver}
C.~{Chen}, D.~{Zhang}, X.~{Ma}, B.~{Guo}, L.~{Wang}, Y.~{Wang}, and E.~{Sha},
  ``crowddeliver: Planning city-wide package delivery paths leveraging the
  crowd of taxis,'' {\em IEEE Transactions on Intelligent Transportation
  Systems}, vol.~18, no.~6, pp.~1478--1496, 2017.

\bibitem{CIT}
C.~{Chen}, Z.~{Wang}, and D.~{Zhang}, ``Sending more with less: Crowdsourcing
  integrated transportation as a new form of citywide passenger–package
  delivery system,'' {\em IT Professional}, vol.~22, no.~1, pp.~56--62, 2020.

\bibitem{nguyen2015people}
N.-Q. Nguyen, N.-V.-D. Nghiem, P.-T. Do, K.-T. LE, M.-S. Nguyen, and N.~Mukai,
  ``People and parcels sharing a taxi for tokyo city,'' in {\em Proceedings of
  the Sixth International Symposium on Information and Communication
  Technology}, pp.~90--97, 2015.

\bibitem{ghilas2013integrating}
V.~Ghilas, E.~Demir, and T.~Van~Woensel, ``Integrating passenger and freight
  transportation: Model formulation and insights,'' {\em Proceedings of the
  2013 Beta Working Papers (WP)}, vol.~441, 2013.

\bibitem{masson2017optimization}
R.~Masson, A.~Trentini, F.~Lehu{\'e}d{\'e}, N.~Malh{\'e}n{\'e}, O.~P{\'e}ton,
  and H.~Tlahig, ``Optimization of a city logistics transportation system with
  mixed passengers and goods,'' {\em EURO Journal on Transportation and
  Logistics}, vol.~6, no.~1, pp.~81--109, 2017.

\bibitem{fatnassi2015planning}
E.~Fatnassi, J.~Chaouachi, and W.~Klibi, ``Planning and operating a shared
  goods and passengers on-demand rapid transit system for sustainable
  city-logistics,'' {\em Transportation Research Part B: Methodological},
  vol.~81, pp.~440--460, 2015.

\bibitem{chen2018multi}
W.~Chen, M.~Mes, and M.~Schutten, ``Multi-hop driver-parcel matching problem
  with time windows,'' {\em Flexible services and manufacturing journal},
  vol.~30, no.~3, pp.~517--553, 2018.

\bibitem{DBLP}
K.~Arulkumaran, M.~P. Deisenroth, M.~Brundage, and A.~A. Bharath, ``A brief
  survey of deep reinforcement learning,'' {\em arXiv preprint
  arXiv:1708.05866}, 2017.

\bibitem{taxi2018limousine}
N.~Taxi, ``Limousine commission-trip record data,'' 2018.

\bibitem{schultz2018deep}
L.~Schultz and V.~Sokolov, ``Deep reinforcement learning for dynamic urban
  transportation problems,'' {\em arXiv preprint arXiv:1806.05310}, 2018.

\bibitem{oda2018movi}
T.~Oda and C.~Joe-Wong, ``Movi: A model-free approach to dynamic fleet
  management,'' in {\em IEEE INFOCOM 2018-IEEE Conference on Computer
  Communications}, pp.~2708--2716, IEEE, 2018.

\bibitem{al2019deeppool}
A.~O. Al-Abbasi, A.~Ghosh, and V.~Aggarwal, ``Deeppool: Distributed model-free
  algorithm for ride-sharing using deep reinforcement learning,'' {\em IEEE
  Transactions on Intelligent Transportation Systems}, vol.~20, no.~12,
  pp.~4714--4727, 2019.

\bibitem{singh2019distributed}
A.~Singh, A.~Alabbasi, and V.~Aggarwal, ``A distributed model-free algorithm
  for multi-hop ride-sharing using deep reinforcement learning,'' {\em arXiv
  preprint arXiv:1910.14002}, 2019.

\bibitem{shaheen2018shared}
S.~Shaheen, ``Shared mobility: the potential of ridehailing and pooling,'' in
  {\em Three Revolutions}, pp.~55--76, Springer, 2018.

\bibitem{DDQN_arxiv}
H.~v. Hasselt, A.~Guez, and D.~Silver, ``Deep reinforcement learning with
  double q-learning,'' in {\em Proceedings of the Thirtieth AAAI Conference on
  Artificial Intelligence}, pp.~2094--2100, 2016.

\bibitem{gasprices}
AAA, ``Us state gas prices.''
  \url{https://gasprices.aaa.com/state-gas-price-averages/}, 2020.
\newblock Accessed: 2020-01-30.

\end{thebibliography}
